\documentclass[10pt,journal,compsoc]{IEEEtran}

\usepackage{array}
\usepackage{mdwmath}
\usepackage[cmex10]{amsmath}
\usepackage{amssymb}
\usepackage[pdftex]{graphicx}
\DeclareGraphicsExtensions{.pdf}
\usepackage{float}
\usepackage{stfloats}
\usepackage{booktabs}
\usepackage{multirow}
\usepackage{multicol}
\usepackage{threeparttable}
\usepackage{graphics}
\usepackage{epstopdf}
\usepackage{color}
\usepackage{steinmetz}

\ifCLASSOPTIONcompsoc
  \usepackage[nocompress]{cite}
\else
  \usepackage{cite}
\fi

\ifCLASSINFOpdf
\else
\fi
\begin{document}
%
\title{TraInterSim: Adaptive and Planning-Aware Hybrid-Driven Traffic Intersection Simulation}

\author{Pei Lv, Xinming Pei, Xinyu Ren, Yuzhen Zhang, Chaochao Li, and Mingliang Xu
\IEEEcompsocitemizethanks{\IEEEcompsocthanksitem Pei Lv, Xinming Pei, Xinyu Ren, Yuzhen Zhang, Chaochao Li, and Mingliang Xu are with the  School of Computer and Artificial Intelligence, Zhengzhou University, Zhengzhou, China.\protect\\
E-mail: ielvpei@zzu.edu.cn, kevinpeixinming@foxmail.com, \{renxinyu, zyzzhang\}@gs.zzu.edu.cn, \{ieccli, iexumingliang\}@zzu.edu.cn}
\thanks{Manuscript received April 19, 2005; revised August 26, 2015.}}

%
%

\markboth{Journal of \LaTeX\ Class Files,~Vol.~14, No.~8, August~2015}%
{Shell \MakeLowercase{\textit{et al.}}: Bare Advanced Demo of IEEEtran.cls for IEEE Computer Society Journals}
%



\IEEEtitleabstractindextext{%
\begin{abstract}
Traffic intersections are important scenes that can be seen almost everywhere in the traffic system. Currently, most simulation methods perform well at highways and urban traffic networks. In intersection scenarios, the challenge lies in the lack of clearly defined lanes, where agents with various motion plannings converge in the central area from different directions. Traditional model-based methods are difficult to drive agents to move realistically at intersections without enough predefined lanes, while data-driven methods often require a large amount of high-quality input data. Simultaneously, tedious parameter tuning is inevitable involved to obtain the desired simulation results. In this paper, we present a novel adaptive and planning-aware hybrid-driven method (TraInterSim) to simulate traffic intersection scenarios. Our hybrid-driven method combines an optimization-based data-driven scheme with a velocity continuity model. It guides the agent's movements using real-world data and can generate those behaviors not present in the input data. Our optimization method fully considers velocity continuity, desired speed, direction guidance, and planning-aware collision avoidance. Agents can perceive others' motion plannings and relative distances to avoid possible collisions. To preserve the individual flexibility of different agents, the parameters in our method are automatically adjusted during the simulation. TraInterSim can generate realistic behaviors of heterogeneous agents in different traffic intersection scenarios in interactive rates. Through extensive experiments as well as user studies, we validate the effectiveness and rationality of the proposed simulation method.
\end{abstract}

}

\maketitle

\IEEEdisplaynontitleabstractindextext

%
\IEEEpeerreviewmaketitle

\ifCLASSOPTIONcompsoc
\IEEEraisesectionheading{\section{Introduction}\label{sec:introduction}}
\else
\section{Introduction}
\label{sec:introduction}
\fi

\IEEEPARstart{T}{raffic} simulation has been widely used in many typical applications such as autonomous driving, urban planning, and computer games. Traffic intersection is an important part of traffic system, however, existing works mainly focus on freeways or city-scale traffic  \cite{ref8}, \cite{ref10}, \cite{ref24}, \cite{ref53}. The intersection scenario contains traffic lights and different types of agents from various directions. In such complex environments, it is challenging to efficiently simulate plausible traffic behaviors of different types of agents.

\begin{figure}
    \begin{minipage}{0.48\linewidth}{}
\vspace{3pt}
\centerline{\includegraphics[totalheight=2.97cm]{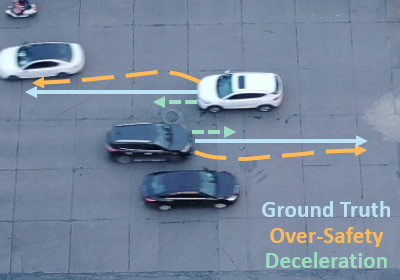}}
\centerline{(a)}
\end{minipage}
\begin{minipage}{0.48\linewidth}
\vspace{3pt}
\centerline{\includegraphics[totalheight=2.97cm]{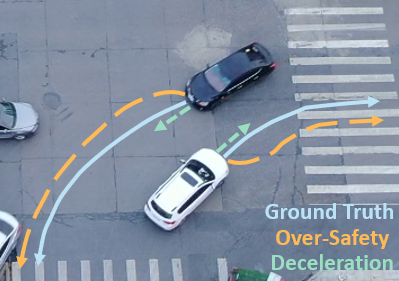}}
\centerline{(b)}
\end{minipage}

\caption{Vehicles at intersections that attempt to avoid collisions only based on the safe distance without considering the motion planning may lead to implausible behavior, such as excessive safety (blue) or unnecessary deceleration (green). (a) Vehicles running straight from opposite directions. (b) Vehicles turning right and left from different directions.}
\label{fig. 1}
\end{figure}

In real world intersections, the motion planning (i.e., turn left/right, go straight) of vehicles plays an important role in determining their movements, which cannot be ignored. Many methods have been proposed to imitate the movement and interaction behavior of agents in crowd simulation and traffic simulation, such as force-based methods \cite{ref1}, \cite{ref2}, velocity-based methods \cite{ref3}, \cite{ref4}, vision-based methods \cite{ref5}, etc. Whereas, these methods are difficult to be directly applied to simulate traffic intersection scenarios without well-defined lanes in the central area. It is easy for agents to react implausibly to avoid possible collisions if the agents' motion plannings are ignored (Fig. \ref{fig. 1}). The traffic simulator \cite{ref6} and \cite{ref7} pre-define lanes and vehicles drive along queues by the car-following model. However, their simulation  results are too regular and inflexible.

Recently, with the development of data acquisition technology, data-driven methods have gradually gained more and more attention. Researchers use ground-truth data (e.g., sensor data \cite{ref10}, \cite{ref11}, GPS data \cite{ref12}, \cite{ref13}) to synthesize trajectories or learn the motion features of agents to reconstruct the traffic flow. Traffic simulation can also use advanced trajectory prediction techniques \cite{ref55}, \cite{ref56}, \cite{ref57}, \cite{ref58} to reproduce traffic flow at signalized intersections. These works can make up the shortage of model-based methods which are difficult to formulate a set of perfect rules to control the motion of agents. However, their results are highly dependent on the quality and quantity of input data. If the input data is flawed (e.g., few data, non-uniform data distribution, noise artifacts), data-driven methods are difficult to obtain satisfactory results. 

On the other hand, most existing methods introduce plenty of parameters to better control the agents. These parameters need to be adjusted iteratively to achieve desired simulation results. Different kinds of agents are mixed in the traffic intersection scenario, which further increases the number of parameters. Some works use empirical data to automatically estimate the parameters of existing simulation methods \cite{ref14}, \cite{ref15}, \cite{ref16}, \cite{ref17}. The results of the above work are also limited by the input data. Moreover, the homogeneous agents in these works using same parameters will inevitably generate similar movement patterns. And the individual flexibility of agents in real-world intersection cannot be efficiently reflected.

In order to overcome above limitations, we propose a novel adaptive and planing-aware hybrid-driven simulation method for traffic intersection scenarios. Inspired by \cite{ref9}, our method combines an optimization-based data-driven approach with a velocity continuity model to generate plausible simulation results even the input dataset is flawed. Specifically, our optimization function fully considers velocity continuity, desired speed, direction guidance, and collision avoidance. In collision avoidance, we integrate not only relative distances between agents but also their motion plannings. In particular, we do not directly update the agent with the velocity in the input dataset that minimizes the optimization function. Instead, we design indicators to decide whether this velocity can be used to update the movement of an agent. If this velocity is unacceptable, we will feed back a set of supplementary candidate velocities based on the agent's current velocity through the continuity of velocity into the optimization process. To reduce tedious parameter tuning and allow agents to have more flexible trajectories, we dynamically estimate the parameters of the proposed method for each agent during the simulation. The input in our method can be different types of trajectory data collected by various devices. Afterwards, we convert those trajectories in the input dataset into the candidate velocity dataset. To improve the quality of these initial trajectories, we further design a curve fitting method to optimize the input trajectories and segment them according to their spatial distribution.

\begin{figure*}[b]
\begin{minipage}{0.32\linewidth}{}
\vspace{3pt}
\centerline{\includegraphics[totalheight=2.77cm]{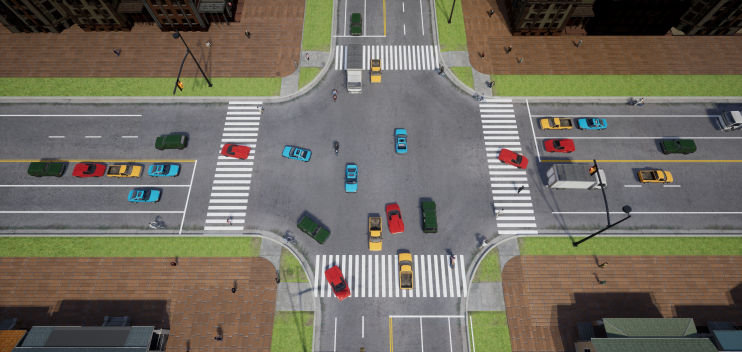}}
\centerline{(a)}
\end{minipage}
\begin{minipage}{0.32\linewidth}
\vspace{3pt}
\centerline{\includegraphics[totalheight=2.77cm]{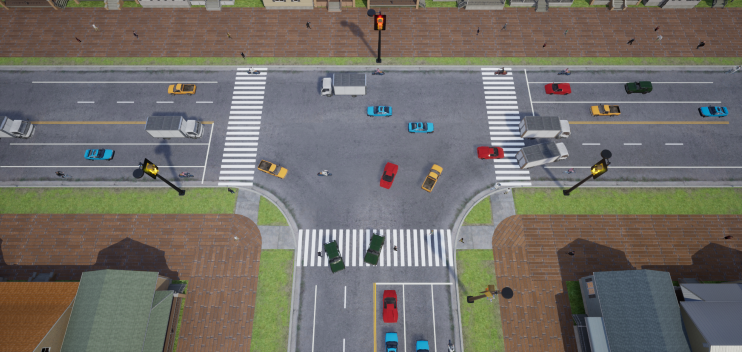}}
\centerline{(b)}
\end{minipage}
\begin{minipage}{0.32\linewidth}
\vspace{3pt}
\centerline{\includegraphics[totalheight=2.77cm, width = 5.83cm]{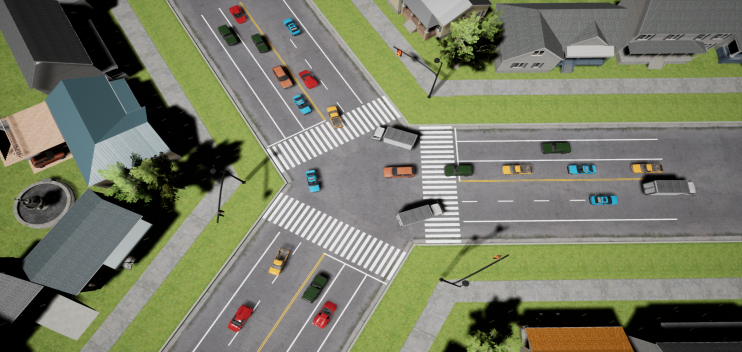}}
\centerline{(c)}
\end{minipage}

\caption{Examples of our simulation in different traffic intersection scenarios with a bird's-eye view. (a) Heterogeneous agents moving in a crossroad. (b) Heterogeneous agents moving in a T-junction. (c) Vehicles driving in a Y-junction.}
\label{fig. 2}
\end{figure*}

The main contributions of our work can be summarized as follows:
\begin{itemize}
    \item We propose a hybrid-driven method to control the movement of agents in traffic intersection scenarios. It combines an optimization-based data-driven scheme with a velocity continuity model, which can effectively drive agents to move realistically even if the input data is defective. 
    \item We introduce a novel collision avoidance energy function integrating the motion planning of agents. Different plausible behaviors of agents can be simulated at traffic intersections without pre-defined lanes.
    \item We present an adaptive parameter adjustment mechanism to dynamically optimize the parameters of our hybrid-driven method during the simulation. It can reflect the flexibility of agent motion and the diverse responses of agents to traffic lights efficiently.
\end{itemize}

Through extensive experiments and user studies, it is demonstrated that our method can generate more realistic traffic simulation results than existing methods. Fig. \ref{fig. 2} shows some snapshots generated by the proposed method on different traffic intersection scenarios.

\section{Related Work}
In this section, we first overview the prior traffic simulation methods, including rule-based methods and data-driven methods. Then we review the works related to multi-agent navigation and parameter estimation.

\subsection{Traffic Simulation}
Early work mainly used rule-based methods to control the motion of agents. In computer graphics community, they can be divided into two categories based on the level of simulation details. Microscopic methods mainly focus on the details of individual behaviors, vehicles are treated as discrete autonomous agents that satisfy certain constraints \cite{ref18}, \cite{ref19}, \cite{ref20}, \cite{ref21}. The Intelligent Driver Model (IDM) \cite{ref22} and the lane-changing model \cite{ref23} are noteworthy, which have been improved and expanded by many subsequent works \cite{ref24}, \cite{ref25}, \cite{ref26}. Macroscopic methods focus on the overall traffic trend, where traffic flows are viewed as continuous dynamics such as fluid or gas. The regulations of vehicle motion are described by solving partial differential equations of fluid dynamics \cite{ref27}, \cite{ref28}, \cite{ref29}, \cite{ref30}, \cite{ref31}. However, these techniques mainly focus on the decision-making process of vehicles on freeways (i.e., acceleration/deceleration or lane-changing). It is difficult to formulate a complete set of rules for agents to match the diverse behaviors in real-world intersections. 

Recently, with the development of data acquisition techniques, more attention has been paid to data-driven simulation methods. Researchers extract information from ground-truth data to reconstruct traffic flow. Spatio-temporal data from in-road sensors are used to reconstruct realistic traffic flow \cite{ref10}, \cite{ref11}. Li et al. \cite{ref12}, \cite{ref13} estimate traffic dynamics using sparse GPS tracks. The specific driving characteristics of drivers are learned from videos in \cite{ref36}. Bi et al. \cite{ref38} simulate the lane-changing process using the vehicle trajectory data. They use randomized forest and back-propagation neural network to learn lane-changing characteristics. Chao et al. \cite{ref8} propose a novel data-driven approach using texture synthesis. They use trajectory data to calculate the velocity in each frame that minimizes the traffic texture energy. 
Trajectory prediction technique can also be used as a novel data-driven scheme. Skarkar et al. \cite{ref55} and Huang et al. \cite{ref56} use video and multiple sensors data to learn the behavior of agents at intersections to predict their trajectories. Zhang et al. \cite{ref50} proposed a trajectory prediction method for intersection scenarios considering the impact of traffic lights on vehicles. Based on an in-house collected intersectional traffic dataset, Bi et al. \cite{ref37} propose a deep learning-based editable simulation framework combined with trajectory prediction.
Data-driven methods can generate more precise traffic flow, but their simulation results are highly correlated with the input data. Our method combines data-driven and model-based methods to plausibly guide the agents' movement even the input data is flawed.

\subsection{Multi-Agent Navigation}
There are many researches involving multi-agent path navigation in traffic simulation and crowd simulation. Local navigation is the most popular research area in the microscopic paradigm, especially collision avoidance between agents \cite{ref39}. Many approaches have been proposed to simulate the interactions between agents by modeling natural human characteristics. Helbing et al. \cite{ref40}, \cite{ref41} first propose the social force model (SFM), assuming that the agent is subjected to the resultant "forces" of the environment and other agents, which drives the agent's movement. Han et al. \cite{ref42} and Chao et al. \cite{ref1} use the force-based method to simulate mixed traffic on freeways. Velocity-based models employ the concept of prediction, where the agent selects the best option from many candidate velocities \cite{ref3}, \cite{ref4}. Karamouzas et al. \cite{ref43} combine force-based methods to provide guaranteed collision-free motion. Ren et al.\cite{ref9} formulate the motion decision of heterogeneous agents as an optimization problem. They construct a velocity dataset, and the agent chooses the velocity that minimizes the energy function. Furthermore, Ond$\rm \check{r}$ej et al. \cite{ref5} propose a vision-based method, each agent makes decisions based on the information rendered onto the virtual retina. To further simulate the movement of real-world pedestrians, human psychology is also incorporated into the crowd simulation \cite{ref44}, \cite{ref45}, \cite{ref46}. These works have achieved collision-free motion, however, they cannot be directly applied to intersectional traffic simulation due to the lack of pre-defined lanes. Our planning-aware collision avoidance scheme injects agents' motion plannings, which can be integrated with previous work to simulate intersection scenarios.

\subsection{Parameter Estimation}
Most of existing simulation methods use plenty of parameters to control the details of the simulation, and the choice of parameters can significantly affect the simulation results which require empirical iterative tuning to find the best combination of parameters. Researchers refer to real-world data for parameter estimation by searching for a set of parameters to minimize the evaluation function. Wolinski et al. \cite{ref15} propose several general evaluation metrics for crowd simulation algorithms. Berseth et al. \cite{ref14} propose a general method to automatically find the optimal parameters for steering algorithms in crowd simulation. Zhou et al. \cite{ref48} model pedestrian dynamics and then learn the parameters of the model from crowd videos. In traffic simulation, Kesting et al. \cite{ref17} calibrate the parameters of IDM and velocity difference model using genetic algorithm (GA). Yang et al. \cite{ref49} use GA to estimate the parameters of their modified SFM to simulate the intersection with an island work zone. Chao et al. \cite{ref16} employ an adaptive genetic algorithm (AGA) to estimate the parameters of their force-based model. These methods need to extract a fixed set of parameters offline using a large amount of sampled data. Different from them, the parameters of each agent in our method are dynamically adjusted at each time step.

\section{Overview}
\begin{figure*}[t]
    \centering
    \includegraphics[width=18.2cm]{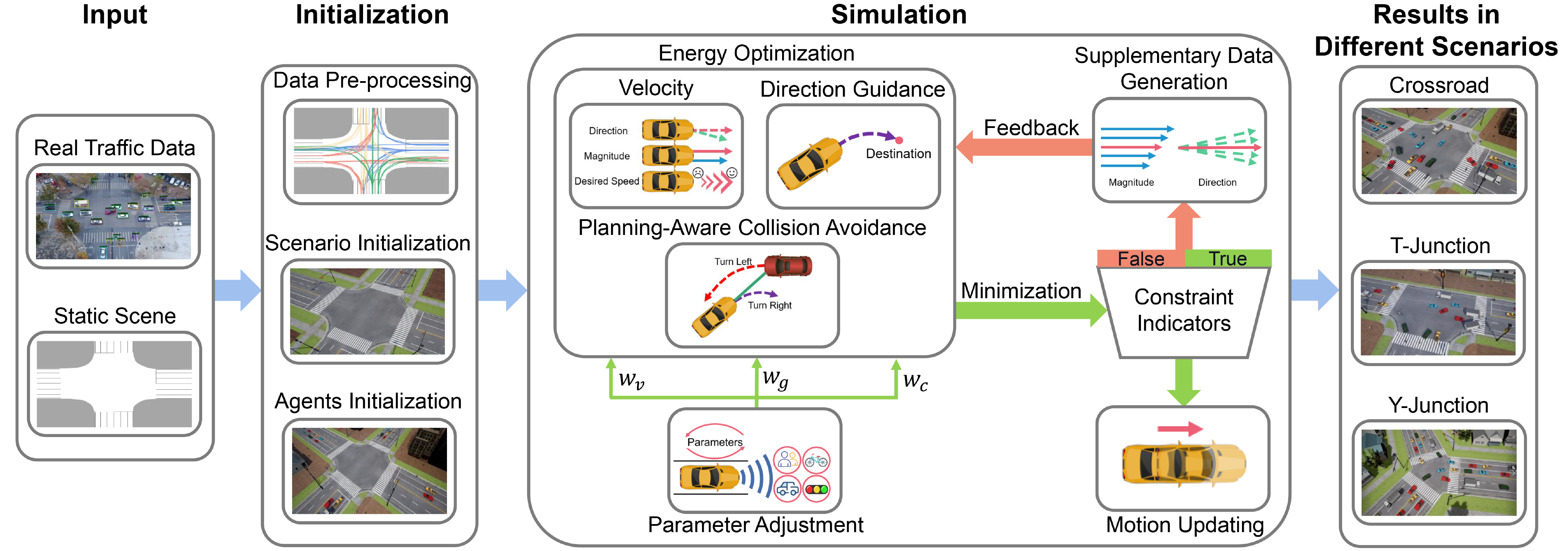}
    \caption{Overview. The input data can be trajectory extracted from video or other data sources that can extract velocity. These velocities are converted into candidate velocity datasets after denoising and division. In the initialization, the static environment and agent information will be set. The parameters in our hybrid-driven method are estimated dynamically at each time step. Our energy function combines the velocity term, direction guidance term, and planning-aware collision avoidance term. Instead of directly use the optimal velocity that numerically minimizes the energy function to updating the agent's state, we set up constraint indicators to determine whether this velocity is acceptable. Our method can simulate heterogeneous agents in different traffic intersection scenarios.}
    \label{fig. 3}
\end{figure*}

Our adaptive and planning-aware hybrid-driven method can simulate the heterogeneous agents in different traffic intersection scenarios with defective input data and without tedious parameter tuning manually. Fig. \ref{fig. 3} provides the overview of our method.

We use traffic trajectory dataset as the input of our method. Since noise is inevitable in the input dataset, we design a curve fitting approach to remove them (Section \ref{sec: 4.1}). It can also be used for other dataset with significant noise or lower sampling rates. The denoised trajectories are grouped according to their spatial distribution. Afterwards, they are converted to velocities, which are used to construct candidate velocity datasets in our method (Section \ref{sec:4.2}). Each candidate velocity dataset is sorted by the magnitude of the velocity. The intersection scene consists of static lanes and dynamic traffic lights. Before the simulation starts, we initialize the position and velocity information for each agent.

Inspired by \cite{ref9}, we also treat the movement decision-making progress of the agent at each time step as an optimization problem. Each agent searches the dataset for a velocity to minimizes the energy function (Section \ref{sec:5.1}). The difference with \cite{ref9} is that we do not directly use this velocity to update the agent's state. Instead, we design indicators to confirm that a velocity is acceptable, including energy terms, velocity magnitude, and direction. If the optimal velocity in the constructed dataset exceeds any threshold, our method will repeat the optimization process by supplementing a series of candidate data with the velocity continuity (Section \ref{sec:5.3}). The agent then updates its state with the final result and adds that velocity to its corresponding dataset. Therefore, our method can simulate the behavior that does not exist in the input data, and the dataset is expanded dynamically during simulation. Our energy function comprehensively considers the continuity of velocity, direction guidance, expected velocity, and collision avoidance (Section \ref{sec:5.2}). The agent's motion planning is injected into our collision avoidance energy function to exhibit plausible agent interaction in intersection areas without predefined lanes. The characteristics of different types of agent behavior are reflected by different parameters of energy functions. We allocate a set of initial parameters that agents can move normally in the sparse traffic flow. In order to show the individual flexibility and dynamics of homogeneous agents, at each time step, we update the parameters of each agent based on the agent's environmental information (Section \ref{sec:5.4}).

\section{Data Pre-processing}
In this section, we describe the details of the input data pre-processing. The input of our method comes from trajectory datasets in different scenarios. To improve the quality of simulation results, we denoise the sampling trajectories and then construct the candidate velocity dataset.

\subsection{Trajectory Denoising} \label{sec: 4.1}
\begin{figure}[t]
    \centering
    \includegraphics[totalheight = 6cm]{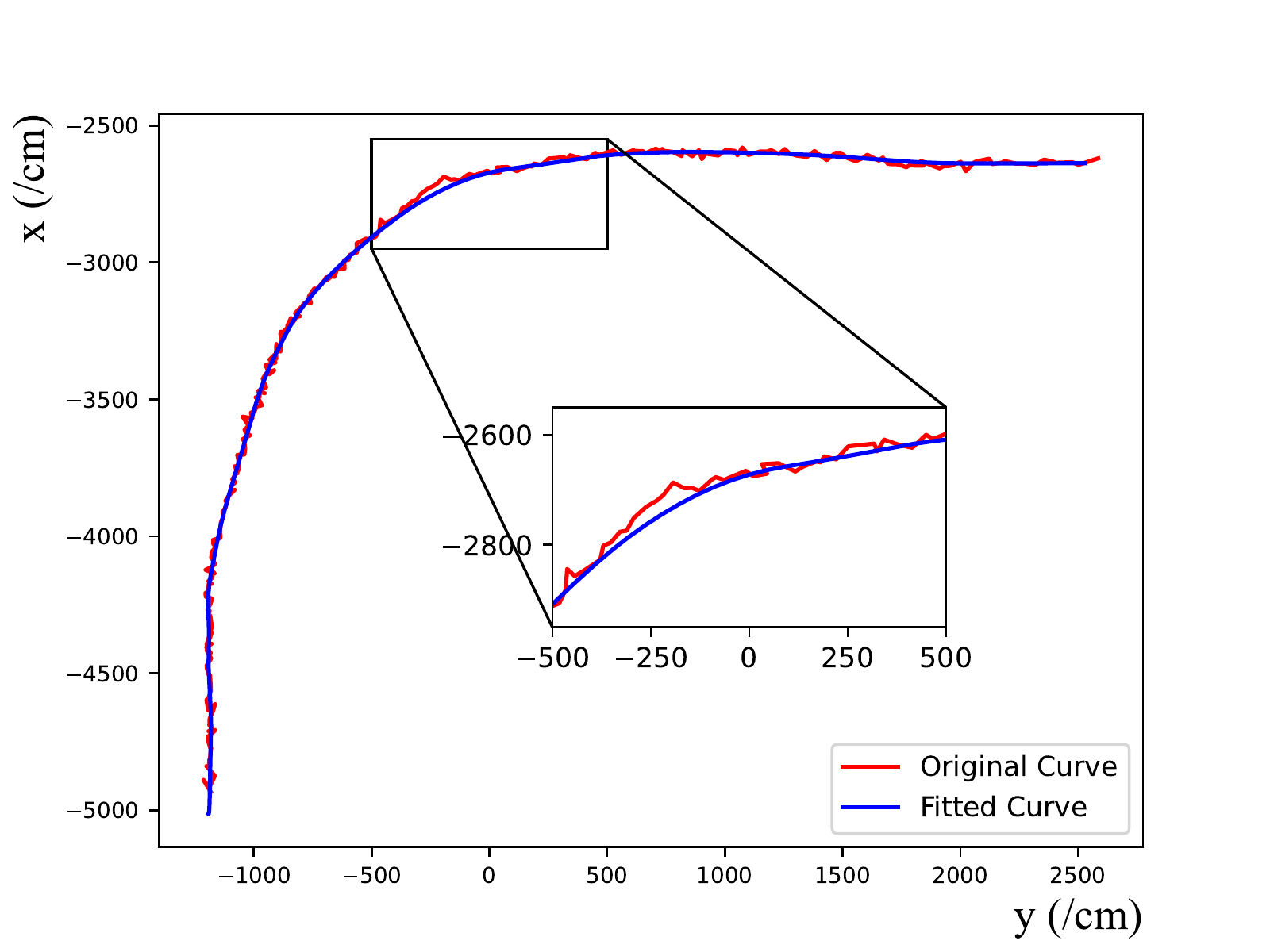}
    \caption{Trajectory Denoising. We use curve fitting to obtain smoother trajectories to eliminate noise in sampled trajectories.}
    \label{fig.4}
\end{figure}

Noise is inevitably introduced in traffic data collection and processing. A large amount of noise data may cause agents to be unable to find velocities suitable for updating their own motion state. We use curve fitting to reduce noise in sampled trajectories. Trajectory curves are interpreted as parametric curves taking frame $t$ as the parameter. The $\Gamma^{S}$ specifies the set of sampling trajectories. The representation for the $j$th trajectory is:
\begin{equation}
    \Gamma^{S}_{j}(t)=
    \begin{cases}
    x^{S}_{j}(t)\\
    y^{S}_{j}(t)
    \end{cases}
     t \geq 0,
\end{equation}
where $x^{S}_{j}(t)$ and $y^{S}_{j}(t)$ are the coordinate of the $X$-axis and $Y$-axis at frame $t$, respectively. We use B-spline to maintain the locality of the curve, so as to prevent some sampling points with large errors from affecting the whole curve. We fit the $X$-axis direction and $Y$-axis direction respectively, and they have the same loss function. For the $X$-axis direction, the equation is defined as follows:
\begin{equation}
    L^{X}=w_{1}L^{X}_{p}+w_{2}L^{X}_{v}+w_{3}L^{X}_{c}+w_{4}L^{XY},
\end{equation}
where $L^{X}_{p}$ is the loss function for the distance from the fitting point to the simpling point, $L^{X}_{v}$ and $L^{X}_{c}$ are the loss functions for the continuity of velocity in magnitude and direction, $L^{XY}$ is the loss function that ensures the continuity of the trajectory after the combination of $X$-axis direction and $Y$-axis direction, and $w_{1}$, $w_{2}$, $w_{3}$, $w_{4}$ are the corresponding weights. The loss functions $L^{X}_{p}$, $L^{X}_{v}$, $L^{X}_{c}$ and $L^{XY}$ are weighted with $w_{1} = 0.7$, $w_{2}=30$, $w_{3}=20$ and $w_{4}=30$, respectively. The results of trajectory denoising are shown in Fig. \ref{fig.4}.

The term $L^{X}_{p}$ is designed to ensure the result having a similar shape to the sample curve:
\begin{equation}
    L^{X}_{p}=\left\Vert x_{j}(t)-x^{S}_{j}(t) \right\Vert_{2},
\end{equation}
where $x_{j}(t)$ is the $X$-coordinate of the fitted curve $j$ at the $t$th frame.

Agent will not change its motion state frequently or significantly in consecutive time frames at intersections, so we introduce the term $L^{X}_{v}$ and $L^{X}_{c}$ to ensure the motion continuity. They are defined in Eq. \eqref{eq.4} and \eqref{eq.5}, respectively
\begin{equation}
    L^{X}_{v}=\left\Vert x_{j}(t)-x_{j}(t-1) \right\Vert_{2},
    \label{eq.4}
\end{equation}
\begin{equation}
    L^{X}_{c}=\left\Vert x_{j}(t)-\frac{1}{2K}\sum\limits_{k=1}^{K}[x_{j}(t-k)+x_{j}(t+k)] \right\Vert_{2}.
    \label{eq.5}
\end{equation}
Among them, $x_{j}(t-1)$ denote the $X$-coordinate of the fitted curve $j$ at previous frame, and $K$ denotes the consecutive adjacent frames.

The above terms can only constrain the current direction, but the combination of two directions may break the continuity of velocity. The term $L^{XY}$ is used to keep the motion continuity after combining $x_{j}(t)$ and $y_{j}(t)$:
\begin{equation}
    L^{XY}=L^{XY}_{v}+L^{XY}_{d},
    \label{eq.6}
\end{equation}
where $L^{XY}_{v}=\left\Vert \Vert {\bf{v}}_{j,t} \Vert-\Vert {\bf{v}}_{j,t-1} \Vert \right\Vert_{2}$ is for the continuity of velocity magnitude, and $L^{XY}_{d}=\left\Vert \hat{\bf{v}}_{j,t}-\hat{\bf{v}}_{j,t-1} \right\Vert_{2}$ is for the continuity of direction.

\subsection{Candidate Velocity Dataset Construction} \label{sec:4.2}
\begin{figure}[t]
    \centering
    \includegraphics[height=5.cm, width = 8.2cm]{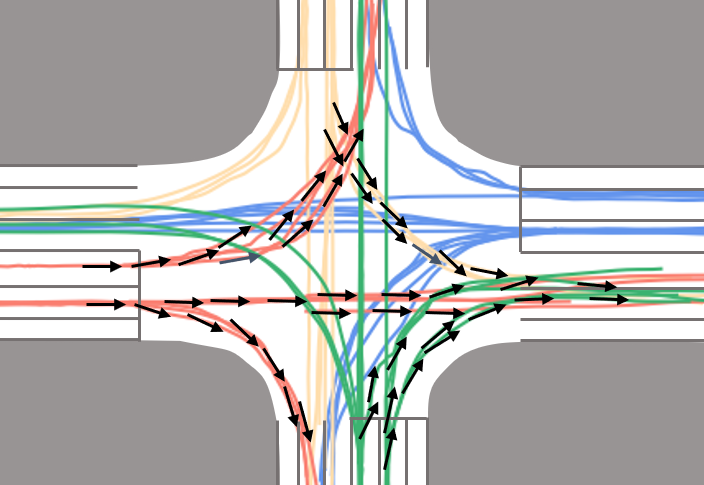}
    \caption{Dataset division. The trajectories of vehicles entering the intersection from the same lane are represented by the same color. Vehicles entering the intersection from the right lane (blue) rarely update their states by selecting the velocities in some trajectory segments (black).}
    \label{fig.5}
\end{figure}

Agents with similar starting points and destinations may have similar movement routes. Therefore, when computing the energy function, most of the candidate velocities required by an agent to update its state come from the trajectories of its own lane. To optimize the solution space, we reduce many velocities which the agent would not choose by dividing the dataset, which ensures there are more acceptable choices for the agent in the limited range of candidates. Taking the vehicles in the intersection as an example, all trajectories correspond to the four lanes entering the crossroad are divided into four groups. Vehicles entering the central area of the intersection from the same lane are grouped together (Fig. \ref{fig.5}). We convert trajectories into velocity $v_{j}(t)=({\Gamma_{j}(t)-\Gamma_{j}(t-1)}) / {\Delta t}$ to obtain candidate velocity dataset $D=D_{1}\cup D_{2}\cup D_{3}\cup D_{4}$, where $\Gamma$ is the trajectory set after denoising, $\Delta t$ is the time difference between two consecutive frames, and $D_{1}$, $D_{2}$, $D_{3}$, $D_{4}$ store the velocities of vehicles entering the intersection from the same lane, respectively. All candidate velocity datasets are sorted by the magnitude of velocity. Although there are some potentially useful velocities excluded that may reduce the diversity of the input data, our hybrid- driven method (Section \ref{sec:5.3}) can attenuate for this. We construct candidate velocity dataset for pedestrians and bicycles in the same scheme. At the beginning of the traffic simulation, the agent selects the corresponding candidate velocity dataset according to its departure lane.

\section{Traffic Intersection Simulation}
In this section, we describe the details of our adaptive and planning-aware hybrid-driven method for traffic intersection simulation. 


\subsection{Problem Formulation} \label{sec:5.1}
Given the initial velocity, desired speed, the point of departure, and destination of agents in the scenario, we aim to simulate plausible trajectories of these agents. The initial candidate velocity dataset is denoted as $D=\bigcup_{n} {\bf{v}}_{n}$, where ${\bf{v}}_{n}$ is the velocity at frame $t_{n}$ converted from trajectory data. Our supplementary dataset is denoted as $D_s=D\bigcup S({\bf{v}}_{i,t-1})$, where $S()$ is our data supplement function. $w$ are the parameters of the energy function. We use $s_{i,t}=[{\bf{p}}_{i,t},{\bf{v}}_{i,t},{\bf{g}}_{i,t},e_{i,t}]$ to specify the state of agent $i$ at time $t$, where ${\bf{p}}_{i,t} \in{\mathbb{R}^2} $ denote the position, ${\bf{v}}_{i,t} \in{\mathbb{R}^2}$ denote the velocity, ${\bf{g}}_{i,t} \in{\mathbb{R}^2}$ and $e_{i,t}\in{\mathbb{R}}$ respectively denote the guidance direction and expected speed. Inspired by \cite{ref9}, we also use optimization-based scheme. The agent searches the velocity to minimize the energy function to update motion state. The difference is that we adjust the parameters for each agent at the beginning of each time step, and our candidate velocities are not only from the input dataset, but also generated by velocity continuity model.
The formula for our method to update the state of agent is
\begin{equation}
    \begin{aligned}
    {{\bf{v}}_{i,t}}&=\mathop {\arg \min }\limits _{ {\bf {v}} \in{D_s}} E({w_{i,t}}, {\bf{v}},s_{i,t-1}), 
    \\ {{\bf{p}}_{i,t}}&={{\bf{p}}_{i,t-1}}+{{\bf{v}}_{i,t}\Delta t},
\end{aligned}
\end{equation}
where $E({w_{i,t}}, {\bf{v}},s_{i,t-1})$ is the energy function, $w_{i,t}$ are parameters of agent $i$ at time $t$., ${{\bf{p}}_{i,t-1}}$ is the position of agent $i$ at time $t-1$, and $\Delta t$ is a time step.

\subsection{Energy Optimization} \label{sec:5.2}
The movement of an agent is constrained by two aspects. On the one hand, it comes from itself: physical limitations, destination, and its expected speed. The other is from its surroundings: neighbors, lanes, and traffic lights. We design the energy function $E$ as follows:
\begin{equation}
    E=w_{v}E_{v}+w_{g}E_{g}+w_{c}E_{c},
    \label{eq.8}
\end{equation}
where $E_{v}$ is the velocity optimization term, $E_{g}$ is the direction guidance optimization term, and $E_{c}$ is the planning-aware collision avoidance optimization term. $w_{v}$, $w_{g}$ and $w_{c}$ represents their weights, respectively. These weights are adaptively adjusted in our method, which will be described in Section \ref{sec:5.4}.

\textbf{\emph{Velocity Energy Optimization Term:}} Due to physical limitations, agents will not change their velocity significantly in a short time. Each agent has an expected speed when the density of a certain area is small, the interaction between agents will be weak, and other external interference factors are basically non-existent. The velocity energy $E_{v}$ is designed to indicate the agent to maintain a limited velocity change while trying to achieve its desired speed:
\begin{equation}
    E_{v}=w_{dir}E^{dir}_{v}+w_{m}E^{m}_{v}+w_{e}E^{e}_{v},
    \label{eq.9}
\end{equation}
where $E^{dir}_{v}$ and $E^{c}_{v}$ represent the continuity of velocity in direction and magnitude, respectively, and their definitions are consistent with Eq. \eqref{eq.6}. We define $E^{e}_{v}=\big\Vert \Vert {\bf{v}} \Vert- {e_{i,t}} \big\Vert_{2}$ to implement the agent to tend to approximate its expected velocity. ${e_{i,t}}$ is the maximum magnitude of velocity desired by the agent $i$ at time $t$.

\textbf{\emph{Direction Guidance Energy Optimization Term:}} We use a guidance direction to control agents to move towards their destination: 
\begin{equation}
    E_{g} = \left\Vert \hat{\bf{v}} - {\bf{g}}_{i,t} \right\Vert_{2}.
\end{equation}
In the central area of an intersection, we define the guidance direction ${\bf{g}}_{i,t}= \frac{{{{\bf{p}}^{goal}_{i}}-{{\bf{p}}_{i,t}}}}{\left\Vert {{{\bf{p}}^{goal}_{i}}-{{\bf{p}}_{i,t}}} \right\Vert}$, where ${{\bf{p}}^{goal}_{i}}$ is the destination coordinate of agent $i$. In the lane area, the guidance direction is parallel to the road.

\textbf{\emph{Planning-Aware Collision Avoidance Energy Optimization Term:}} Collision avoidance between agents is the most attended issue in local navigation decisions. The agent should avoid collisions that may occur after a few time steps, but also conform to behaviors of agents in the real world. We treat collision avoidance energy as an implicit influence of surroundings on the agent. And we ensure that the neighbors with greater influence take a larger proportion of the total energy.
\begin{equation}
    E^{total}_{c}=\frac{\sum\nolimits_{\phi\in{\Phi}}{\left\Vert E_c(i,\phi)\right\Vert^{2}}}{\sum\nolimits_{\phi\in{\Phi}}{E_c(i,\phi)}},
    \label{eq.11}
\end{equation}
where $\Phi$ is the set of neighbors in the perception area of agent $i$ at timestep $t$, $\phi$ is any neighbor belonging to $\Phi$, and ${E_c(i,\phi)}$ is the energy function of agent $i$ interacting with neighbor $\phi$.

\begin{figure}[t]
\begin{minipage}{0.48\linewidth}{}
\vspace{3pt}
\centerline{\includegraphics[totalheight=3.4cm, width=4.2cm]{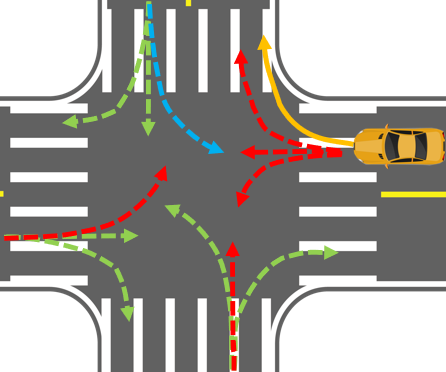}}
\centerline{(a)}
\end{minipage}
\begin{minipage}{0.48\linewidth}
\vspace{3pt}
\centerline{\includegraphics[totalheight=3.4cm]{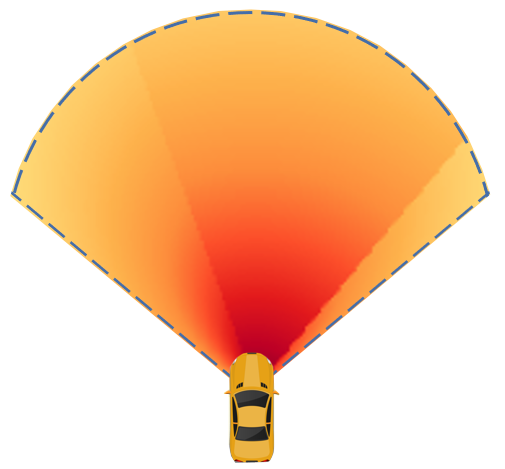}}
\centerline{(b)}
\end{minipage}
    \caption{Motion planning-aware collision avoidance. The impact of motion planning on collision has two aspects. For example, as a right-turning vehicle (yellow) in the crossroad: (a) Whether the interaction occurs depending on the motion plan and location of its neighbors ( high probability(red), small probability(blue) and absolutely impossible(green)). (b) It pays more attention to the agents in front and on the right. In its perception region, darker colors indicate a greater influence.}
    \label{fig.6}
\end{figure}

Due to the lack of well-defined lanes at intersections, agents may make implausible decisions if only relative distances are considered when avoiding collisions (Fig. \ref{fig. 1}). The motion planning of agent cannot be ignored. We define ${E_c(i,\phi)}$ depending on the kind of agents. When agent $i$ and neighbor $\phi$ are homogeneous agents and they are not pedestrians,
\begin{equation}
    E_{c}(i,\phi)= {M(i,\phi)} {f\left( {\xi(i)},\theta_{\phi} \right)}{e^{d_{s}-d_{\phi}}}.
\end{equation}
If $i$ and $\phi$ are both pedestrians or they are heterogeneous agents, the definition is:
\begin{equation}
    E_{c}(i,\phi)={f\left( {\xi(i)},\theta_{\phi} \right)}{e^{d_{s}-d_{\phi}}},
\end{equation}
where ${\xi(i)}$ is the motion planning of agent $i$, $d_{s}$ is the safe distance for agent to avoid colliding with its neighbors, $d_{\phi}$ and $\theta_\phi$ are the distance and the angle between $i$ and $\phi$ after $T=15 \cdot \Delta t$ time steps, respectively. The mask matrix $M(i,\phi)$ is filled with 0, 0.5 and 1.0. Its value is obtained by the motion planning and location of $i$ and $\phi$ to determine whether a collision will happen (Fig. \ref{fig.6}a). The definition of $f\left( {\xi(i)},\theta_{\phi} \right)$ is based on agent $i$'s own motion planning. We assume that the sensitivity of agents to their surroundings is different. For example, when the agent turns right, it is less sensitive to neighbors on the left (Fig. \ref{fig.6}b). We design $f\left( {\xi(i)},\theta_{\phi} \right)$ as a piecewise Gaussian function:
\begin{equation}
\begin{small}
     f\left( {\xi(i)},\theta_{\phi} \right)=
    \begin{cases}
    e^{-\tfrac{\theta^{2}_{\phi}}{2\sigma^{2}_{1}}}, \theta_{\phi} \!\in\! \lbrack -\theta_{\mu_1},\theta_{\mu_2}\rbrack \\ 
    e^{-\tfrac{\theta^{2}_{\phi}}{2\sigma^{2}_{2}}}, \theta_{\phi} \!\in\! \lbrack -50, \pm \theta_{\mu_1}) \!\cup\! (\mp\theta_{\mu_2},50\rbrack
    \end{cases}
\end{small}
\end{equation}
where $\sigma^{2}_{1}$ and $\sigma^{2}_{2}$ are variances used to adjust the sensitivity of the angle, $\theta_\mu$ is the boundary point of the function. $\sigma^{2}_{1}$, $\sigma^{2}_{2}$ and $\theta_\mu$ vary depending on different motion plannings. For vehicles and bicycles, if $\xi(i)$ goes straight, we set $\sigma^{2}_{1}=550$, $\sigma^{2}_{2}=200$, and $\theta_{\mu_1} = \theta_{\mu_2} = 30$; if $\xi(i)$ is turning, we set $\sigma^{2}_{1}=650$, $\sigma^{2}_{2}=250$, $\theta_{\mu_1} = 20$, and $ \theta_{\mu_2} = 40$. For pedestrians, we set $\sigma^{2}_{1}=1200$, $\sigma^{2}_{2}=200$, and $\theta_{\mu_1} = \theta_{\mu_2} = 35$.


\subsection{Supplementary Data Generation} \label{sec:5.3}

The energy functions are not sufficient to provide a complete description of the real-world agent's kinematic characteristics. The method in Section \ref{sec:5.2} is data-driven, and its performance is closely related to the quality of the input data. When the input data is defective, the agent can still find a velocity that minimizes the value of the energy function, but this velocity is not necessarily suitable for updating the agent's motion. For instance, in the extreme case that the input data only has a unique velocity, this velocity will always minimize the value of the energy function. No matter what happens to the surrounding, such as steering and collision avoidance, agents can only select it to move along a straight line.

We set up an indicator $\tau({\bf{v}})$ to measure whether the current numerical optimal speed is behaviorally acceptable:
\begin{equation}
\begin{aligned}
     \tau({\bf{v}}) = &\tau_{m}(\left\Vert \Vert{\bf{v}}_{i,t-1}\Vert-\Vert{\bf{v}}\Vert \right\Vert) \land \tau_{dir}\left(\angle {\bf{\hat{v}}}_{i,t-1} {\bf{\hat{v}}} \right)   \\
     &\land \tau_{g}(w_{g}E_{g}) \land \tau_{c}(w_{c}E_{c}),
\end{aligned}
\end{equation}
where ${\angle \bf{\hat{v}}}_{i,t-1} {\bf{\hat{v}}}$ is the angle between ${\bf{\hat{v}}}_{i,t-1}$ and ${\bf{\hat{v}}}$. Indicator $\tau_{m}$, $\tau_{dir}$, $\tau_{g}$, $\tau_{c}$ are used to measure the magnitude and direction of velocity, guidance direction, and collision avoidance, respectively, which correspond to the energy function in Eq.\eqref{eq.8}. We empirically set $\tau_{m}=\tau_{dir}=1.0$, $\tau_{g} = 2.0$, and $\tau_{c} = 10.0$. 
 
 We assume that the velocity $\bf{v}$ cannot be used to update the state of agent if it does not satisfy the threshold indicator. Then we generate some new candidate data via velocity continuity based on the velocity of the agent in previous frame. The $S_{m}(\bf{v})$ and $S_{d}(\bf{v})$ generate supplementary data in the magnitude and direction of the velocity, respectively:
\begin{equation}
    S_{m}({\bf{v}})\!=\!\left\{ \Vert {\bf{v}}^{*} \Vert  \colon \! \Vert {\bf{v}}^{*} \Vert\!=\! \Vert {\bf{v}} \Vert + pI^{m}, p\!\in\![-\frac{\psi^{m}}{I^{m}},\frac{\psi^{m}}{I^{m}}]\right\}\!,\!
\end{equation}

\begin{equation}
    S_{d}({\bf{v}})=\left\{ \hat{\bf{v}}^{*} \colon \hat{\bf{v}}^{*} = \hat{\bf{v}} +qI^{d},q\in[-\frac{\psi^{d}}{I^{d}},\frac{\psi^{d}}{I^{d}}]\right\}.
\end{equation}
In above equations, $\psi^{m}$ and $\psi^{d}$ denote the range of supplementary velocity in magnitude and direction, $I^{m}$ and $I^{d}$ denote the interval of magnitude and the interval of direction, respectively. Afterwards, we obtain supplementary candidate velocities $\bf{v}^{*}$, and ${\bf{v^{*}}}=\Vert{\bf{v^{*}}}\Vert \hat{\bf{v}}^{*}$. We feed all $\bf{v}^{*}$ back to the energy optimization stage (Section \ref{sec:5.2}) and then select the one that minimizes the energy function as ${\bf{v}}_{i,j}$ to update the state of agent $i$, then add it to $i$'s corresponding candidate velocity dataset.

\subsection{Parameter Adjustment} \label{sec:5.4}
We propose parameter adjustment to reduce tedious adjustment of various parameters and make the agent have more flexible movement patterns. Specifically, at each time step, the parameters of each agent are dynamically computed according to the environment on the basis of a set of benchmark parameters. We alter $w_{v}$ and $w_{g}$ in Eq. \eqref{eq.8}. $E_{c}$ is used to measure the influence between agents, so its weight $w_{c}$ remains unchanged.

\textbf{\emph{Velocity Energy Optimization Term:}} We assume that the weights in $E_v$ are initial values when agents move normally. We adjust parameters in $w_{v}$ in following two cases: ($\rm \romannumeral 1$) agent interaction ($\rm \romannumeral 2$) facing traffic lights. As mentioned in Section \ref{sec:5.2}, $E_c$ is regarded as the influence of neighbors on agents. When there is a possible collision, we adjust the parameters in $E_v$ according to short-term influence:
\begin{equation}
    \begin{aligned}
    w_{dir} &= w^{init}_{dir} + (E^{temp}_{c} / \lambda_{dir}), \\
    w_{m} &= w^{init}_{m} - (E^{temp}_{c} / \lambda_{m}), \\
    w_{e} &= w^{init}_{e} - (E^{temp}_{c} / \lambda_{e}),
\end{aligned}
\end{equation}
where $w^{init}$ denotes the initial parameter and $\lambda$ is used to measure the sensitivity of parameters to neighbors. $E^{temp}_{c}$ is a short-term influence, it differs from $E_{c}$ in Eq. \eqref{eq.11} where it calculates the influence of neighbors on the agent after $T/2$ time steps. When facing an imminent collision, due to the low flexibility of the vehicles, we set them to avoid the collision by decelerating rather than turning suddenly. We empirically set $\lambda_{dir} = \lambda_{m} = \lambda_{e}$=2.0. Pedestrians and bicycles are more agile, they can change direction to avoid possible collisions. We set $\lambda_{dir}=4.5$, $\lambda_{m}=0.5$, $\lambda_{e}=1.5$ for pedestrians, and $\lambda_{dir}=2$, $\lambda_{m}=0.5$, $\lambda_{e}=1.5$ for bicycles.

We treat the traffic light as a linear obstacle containing the state. Agents will gradually stop as they approach the line segment under the red light. In real life, if the traffic light turns red when pedestrians are already inside the intersection area, they will speed up through the crossroad for their safety and to avoid traffic jams. 
Before the simulation starts, we initialize all agents with two desired velocities, $e_1$ and $e_2$ are desired speed for driving normally and accelerating, respectively. When agents need to speed up, we set $e_{i,t}=e_{2}$ and increase $w_{e}=2w^{init}_{e}$ in Eq. \eqref{eq.9}. Some drivers may accelerate to pass under a yellow light. We determine whether there 
is a condition for the agent to accelerate:
\begin{equation}
    \left( \frac{\Vert{\bf{v}}_{i,t}\Vert+e_{2}}{2} t' \right) \geq G,
\end{equation}
where $t'$ is the time when the vehicle accelerates through the stop line, $G$ is the distance between the agent's current position and the point where the entire vehicle completely crosses the stop line. Since the acceleration of a vehicle is not instantaneous, the velocity we use in calculation is the interpolation of the current speed and the desired speed. And we estimate time steps $t'=\varepsilon(s-1)$, $\varepsilon$ is the frame rate, and $s$ is the number of seconds remaining in yellow lights.
\begin{figure}[t]
    \centering
    \includegraphics[width=4.5cm]{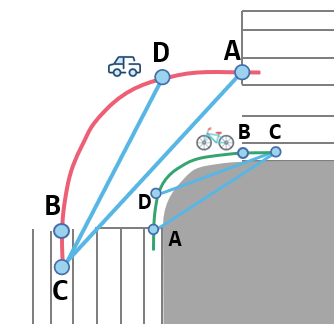}
    \caption{Direction Guidance. The rotation angle of the agent increases when it gets closer to the destination lane. }
    \label{fig.7}
\end{figure}

\textbf{\emph{Direction Guidance Energy Optimization Term:}}
The widths of roads may not be exactly the same, since we do not predefine lanes in the central area of the intersection, the weight $w_{g}$ needs to be repeatedly adjusted to balance the movement of steering vehicles and bicycles in different lanes. According to our observations, the steering range of the vehicle is small when it just drives out of the lane, and gradually increases with the moving process. Based on this, we compute $w_{g}$ :
\begin{equation}
    w_{g} = \frac{dis_{AC}}{dis_{CD}}w^{init}_{g},
\end{equation}
where $A$ and $B$ denote the midpoints of the lane boundaries, $C$ and $D$ denote the destination and current coordinates of the vehicle, $dis_{AC}$ represents the distance between $A$ and $C$, and $dis_{CD}$ represents the distance between $C$ and $D$, respectively (Fig. \ref{fig.7}). We only perform this calculation in the intersection area (i.e., interval $AB$) to avoid $\frac{dis_{AC}}{{dis_{CD}}} \to {\infty}$.

\section{Experimental Results and Analysis}
We conduct experiments on different intersection scenarios to demonstrate our method. Fig. \ref{fig. 2} shows the snapshots of our simulation results in different scenarios, including 4-lane crossroad (Crossroad-1), 2-lane crossroad (Crossroad-2), T-junction, and Y-junction. The corresponding traffic animations are included in the supplementary demo video.

\renewcommand{\arraystretch}{1.4}
\begin{table}[t]
\centering  
\caption{The Initial Parameters in Different Scenarios}  
\label{table1}
\begin{threeparttable}
\setlength\tabcolsep{5pt} 
\begin{tabular}{ccccccc}
\toprule[1pt] 

\multicolumn{2}{c}{Scenario}                                 & $w^{init}_{dir}$ & $w^{init}_{m}$ & $w^{init}_{e}$ & $w^{init}_{g}$ & $w^{init}_{c}$ \\ \hline
\multirow{3}{*}{Crossroad-1}                      & Car        & 1.0  & 1.0  & 1.5  & 1.0  & 1.0  \\
                                                & Pedestrian & 0.5  & 1.0  & 1.5  & 1.3  & 1.0  \\
                                                & Bicycle    & 1.0  & 1.0  & 2.0  & 1.5  & 1.0  \\ 
                                                \hline
\multirow{3}{*}{Crossroad-2}                      & Car        & 1.0  & 1.0  & 1.5  & 1.0  & 1.0  \\
                                                & Pedestrian & 0.5  & 1.0  & 1.5  & 1.3  & 1.0  \\
                                                & Bicycle    & 0.8  & 1.0  & 2.0  & 1.5  & 1.0  \\\hline
\multicolumn{1}{l}{\multirow{3}{*}{T-junction}} & Car        & 1.0  & 1.0  & 1.5  & 1.0  & 1.0  \\
\multicolumn{1}{l}{}                            & Pedestrian & 0.5  & 1.0  & 1.5  & 1.3  & 1.0  \\
\multicolumn{1}{l}{}                            & Bicycle    & 1.0  & 1.0  & 2.0  & 1.5  & 1.0  \\ \hline
Y-junction                                      & Car        & 1.0  & 1.0  & 1.5  & 0.8  & 1.0 \\

\bottomrule[1pt]

\end{tabular}
\emph{Table 1 gives the benchmark parameters of the energy function for different kinds of agents in different scenarios.}
\end{threeparttable}
\end{table}

\begin{table*}[hb]
\centering
    \caption{Runtime Performance in Different Scenarios}
    \label{tab. 2}
\begin{threeparttable}
\setlength\tabcolsep{13pt}
\begin{tabular}{cccccc}
\toprule[1pt] 
Scenario & Types & Agent Number & Dataset & Maximum Time(s/f) & Average Time(s/f) \\ \hline
Crossroad-1       & car/human/bicycle   & 40/30/20  & {[}VTP-TL 2022{]} & 0.04472   & 0.03337   \\ \hline
Crossroad-2       & car/human/bicycle   & 30/10/10  & {[}VTP-TL 2022{]} & 0.04052   & 0.02509   \\ \hline
T-junction        & car/human/bicycle   & 30/30/15  & {[}VTP-TL 2022{]} & 0.04280   & 0.03208   \\ \hline
Y-junction        & car   & 40  & {[}Waymo 2021{]} & 0.04475   & 0.03625 \\
\bottomrule[1pt] 
\end{tabular}
\emph{Table 2 shows the running performance of our method to simulate heterogeneous agents in different traffic intersection scenarios, including the maximum time (seconds per frame) and the average time (seconds per frame). We use different datasets to validate our method.}
\end{threeparttable}
\end{table*}

The input in our method can be trajectory data collected by different devices (various sensors, video tracking, etc.). Our method includes trajectory optimization and supplementary data generation, so small data volume, non-uniform data distribution and noise artifacts are all acceptable. However, since we divide the data according to spatial distribution, it is best to choose trajectories which can distinguish motion plannings. For crossroad scenarios and T-junction scenario, our input samples are selected from VTP-TL Dataset \cite{ref50}, where trajectories are labeled in videos shot by the UAV from different traffic intersection scenes with a 30 fps frame rate. The crossroad scene has four lanes merging into the central area from different directions. We randomly select six cars, four pedestrians and four bicycles from trajectories in each lane. The average duration of each trajectory is 350 frames. For Y-junction scenario, our input are selected from Waymo Open Motion Dataset \cite{ref51}, where trajectories are acquired from sensors installed in the vehicle. Each spatio-temporal segment in this dataset has a duration of 9 seconds with 10hz sampling rate, and we randomly select ten trajectory segments from each lane.

\subsection{Runtime Performance} \label{sec:6.1}
Our approach is implemented with the Unreal Engine in a 64-bit desktop with a 3.80 GHz Intel(R) Core(TM) i7-10700K processor and 16GB memory. We achieve the behavior characteristics of heterogeneous agents by setting different parameters. We use 1.0 as the benchmark to initialize all parameters. Since the parameters in our method are adaptively calculated during the simulation, we only need to manually make minor adjustments to the parameters in $E_v$ according to the motion characteristics of different kinds of agents. Table \ref{table1} lists the parameter values used in our method. We use the scheme mentioned in \cite{ref9} to speed up the computation, where agents only interact with neighbors in adjacent grids and choose velocity from limited search space. Table \ref{tab. 2} shows the runtime performance of our method in different traffic intersection scenarios. 

\begin{figure}[t]
    \centering
    \includegraphics[width=8cm]{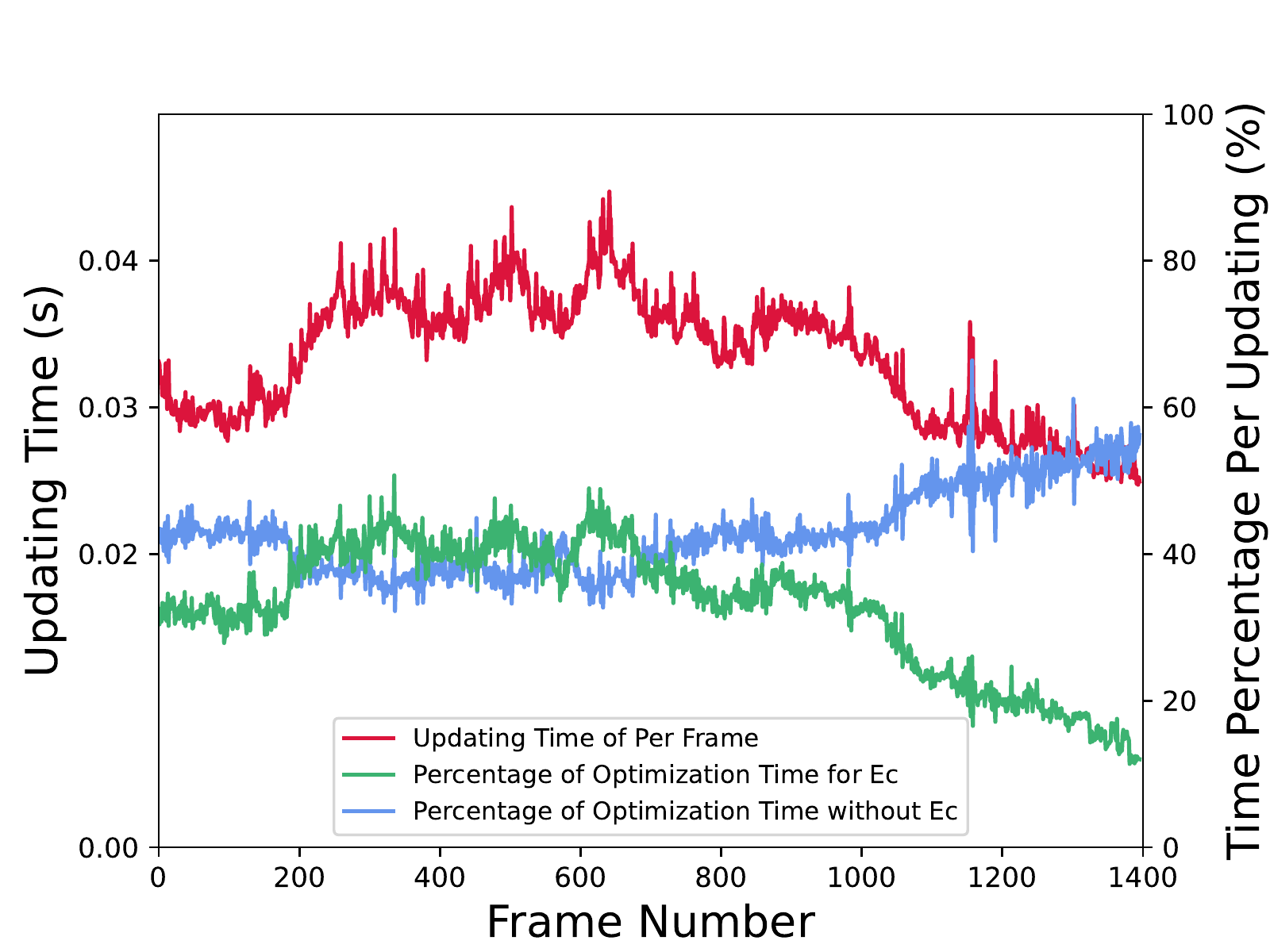}
    \caption{Runtime performance. We analyze the runtime performance of our simulation results in the crossroad scenario. The specific computation time per frame is shown and also the proportion of time spent on computing agent interactions in the total time.}
    \label{fig.8}
\end{figure}

To evaluate the performance of our method in detail, we analyze the specific computation time in the Crossroad-1 scenario (Fig. \ref{fig.8}). We show the computation time per frame and the time to compute agent interactions. Since the size of the solution space in the dataset is the same for each time step, the change in computation time is mainly due to the computational cost of each candidate velocity increased by the interaction between agents. 

\subsection{Comparisons}
\subsubsection{Effectiveness of Supplementary Data} \label{sec:res with SD}
Due to the inherent characteristic stochastic of traffic, direct trajectory comparisons are usually not performed for traffic simulation \cite{ref54}. Vehicles in the traffic intersection scenarios come from different directions and it is difficult to directly calculate the gap between vehicles. To validate the performance of our hybrid-driven method, we compare the distribution of velocity and steering angle. Velocity is the fundamental property for measuring movement, and steering angle is used to measure the performance of turning. 

We verify the effectiveness of supplementary data from both qualitative and quantitative perspectives. In qualitative experiments, we randomly select the trajectories of vehicles with different motion plannings drive past green lights and no neighbor interactions as input data from VTP-TL \cite{ref50}. Using the same input, our simulation results with and without supplementary data are compared with ground truths respectively: ($\rm \romannumeral 1$) comparison with input data. ($\rm \romannumeral 2$) comparison with trajectory segments similar to the input data, where vehicles wait for red lights or interact with neighbors. In the quantitative experiments, we randomly select trajectories as input from lanes entering the intersection from different directions. After constructing the candidate velocity dataset, we randomly delete different proportions of velocity data to simulate the ground truth. The parameters used in the experiments are the same as Section \ref{sec:6.1}. 

\begin{figure}[t]
    \centering
    \includegraphics[height=7.3cm]{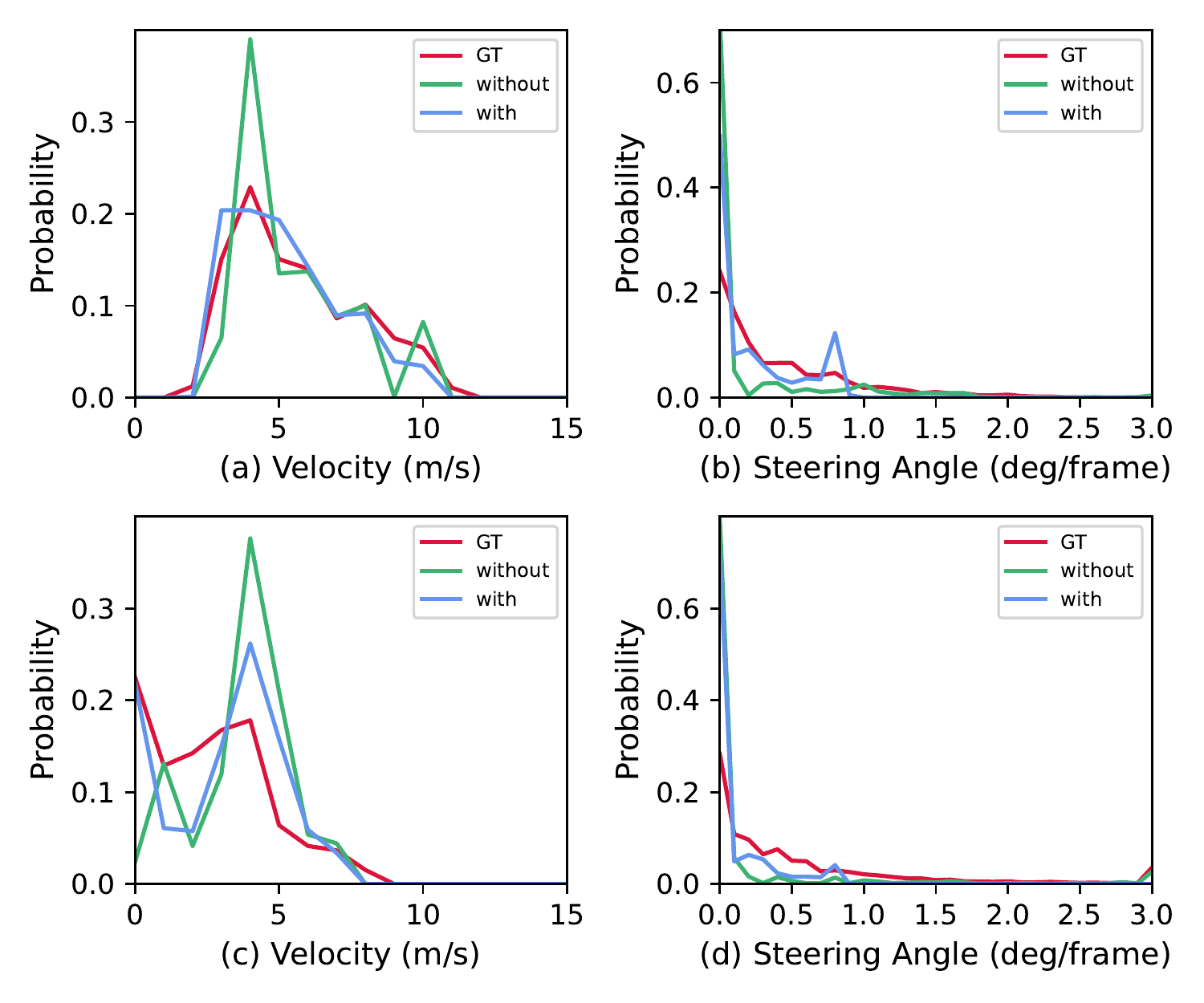}
    \caption{Distributions of velocity and steering angle in the experiments to verify the effectiveness of supplementary data. (a)-(b): Probability distributions when there are no surrounding interactions. (c)-(d): Probability distributions when there are interactions around.}
    \label{fig.9}
\end{figure}

\begin{table}[t]
\caption{Results of Different Percentage Input Data}  
\centering
\label{table remaindata}
\begin{tabular}{cccccc}
\toprule[1pt] 
Input Data & 100\% & 80\% & 60\% & 40\% & 20\% \\ \hline
Velocity             & 0.4112 & 0.4326 & 0.4410 & 0.5078 & 0.5517 \\ 
Steering Angle       & 0.7593 & 0.7992 & 0.8078 & 0.8542 & 0.9593 \\

\bottomrule[1pt] 

\end{tabular}
\end{table}

Fig. \ref{fig.9} shows the experimental results. We calculate the probability of velocity and steering angle in each interval. The probability difference between the ground truth and the simulation result is the sum of the probability differences of all intervals, a lower score indicates a closer approximation to the distribution of the ground truth. In qualitative experiments (Fig. \ref{fig.9}), compared with different ground truths: the velocity differences are 0.2036 : 0.3828 and 0.3902 : 0.7317, the steering angle differences are 0.6675 : 0.9935 and 0.8976 : 1.029. Experimental results show that simulations with supplementary data outperform those without supplementary data. In particular, in the presence of traffic lights and neighbors, vehicles without supplementary data were unable to decelerate to a standstill. In quantitative experiments (Table \ref{table remaindata}), the simulation results using all input data are better than others, simulations using $80\%$  and $60\%$ input data achieve similar results. When only $40\%$ and $20\%$ of the input data are available, the differences between the simulation results and the ground truth is large. The experimental results show that with the decrease of the input data, the distribution differences between the simulation results and ground truth increase gradually.

Both qualitative and quantitative experiments show the effectiveness of supplementary data. There are two main reasons: ($\rm \romannumeral 1$) without the supplementary data, it is impossible to choose velocities that are not included in the input data. ($\rm \romannumeral 2$) limited energy functions and constraints cannot fully model complex real-world environments. Both of these lead the fact that the velocity which minimizes the optimization function is only the numerically optimal solutions, but it not necessarily the most appropriate to update the state of agent.

In steering angles (Fig. \ref{fig.9}b and \ref{fig.9}d), the distribution of our method at zero degree is significantly higher than ground truths. In real world, vehicles have more flexible trajectories. Especially, vehicles whose motion planning is straight usually do not move strictly in a straight line. However, in our method, vehicles have a high probability of choosing the same velocity in successive time steps.

\begin{table}[b]
\centering

\caption{Results with Different Velocity Continuity Space}  
\label{table velSpace}
\setlength\tabcolsep{15pt} 
\begin{tabular}{ccc}
\toprule[1pt] 
$(\psi^{m}, I^{m}; \psi^{d}, I^{d})$ & Velocity & Steering Angle \\ \hline
(1.6, 0.4; 1.2, 0.4) &  0.4932        &       0.7992         \\
(0.8, 0.2; 0.6, 0.2) &    0.5562      &         0.9051       \\
(1.6, 0.4; 0.9, 0.3) &     0.4577     &          0.7931      \\
(0.8, 0.2; 0.9, 0.3) &   0.4592       &          0.7780      \\
(1.0, 0.25; 1.2, 0.4) &     0.4484     &        0.8090        \\
(1.0, 0.25; 0.6, 0.2) &       0.4587   &          0.8217      \\
(1.0, 0.25; 0.9, 0.3) &    0.4112      &          0.7593      \\
\bottomrule[1pt] 
\end{tabular}
\end{table}

The supplementary data of our method is generated by the velocity continuity model. In addition, we carry out ablation experiments in different range of velocity space. The experimental results are shown in Table \ref{table velSpace}. We set a variety of size and direction ranges as well as that of intervals for the velocity continuity model. The experimental results show that the best simulation results can be obtained under $\psi^{m}=1.0$, $I^{m}=0.25$, $\psi^{d}=0.9$, and $I^{d}=0.3$.

\subsubsection{Results with Different Energy Terms}
We perform ablation experiments to verify the effectiveness of each energy term. The distribution differences between simulation results and ground truth in velocity and steering angle are shown in Table \ref{table 3}.
\begin{table}[t]
\centering  
\caption{Results with Different Energy Terms}  
\label{table 3}
\setlength\tabcolsep{13pt} 
\begin{threeparttable}
\begin{tabular}{ccccc}
\toprule[1pt] 
\multirow{2}{*}{$E_v$} & \multirow{2}{*}{$E_g$} & \multirow{2}{*}{$E_c$} & \multicolumn{2}{c}{Metrics} \\ \cline{4-5} 
                       &                        &                        & Velocity  & Steering Angle  \\ \hline
                       & $\surd$                & $\surd$                & 0.4634    & 0.7044          \\ \hline
$\surd$                &                        & $\surd$                & 0.5008    & 0.7814          \\ \hline
$\surd$                & $\surd$                &                        & 0.7794    & 0.6263          \\ \hline
$\surd$                & $\surd$                & $\surd$                & 0.4186    & 0.6286          \\ 
\bottomrule[1pt] 
\end{tabular}
\emph{Table 5 shows the differences of distribution between the simulated results and the ground truth without different energy terms. Agents cannot move without the velocity item $E_{v}$, where we set default value 0.2 for weight $w_{e}$ in this experiment.}
\end{threeparttable}
\end{table}

Experimental results illustrate that the absence of each energy term will significantly affect the simulation results. The results missing any energy term get higher scores, which indicates that their distributions of velocity and steering angle are more different from the ground truth. In the experiment without the velocity term $E_{v}$, we set a default value 0.2 for the weight $w_{e}$, and the agent can move normally but may not reach the desired speed. Therefore, its score is less different from that contains all energy terms. However, there will be a sudden acceleration/deceleration or steering behavior when the agent's speed and direction need to be changed. The distribution difference score of steering angle increases significantly in experiment without the direction guidance term $E_{g}$. The simulation results show that the agents deviate from their destination: ($\rm \romannumeral 1$) The agents whose motion planning is steering can only move in a straight line parallel to the direction of their starting road. ($\rm \romannumeral 2$) The agents with interactive behavior can only swerve to avoid collisions and then continue to move in a straight line. When the agents lack the motion planning-aware collision avoidance term $E_{c}$, the difference in distribution of the velocity increases significantly. The agents cannot avoid an impending collision or stop at a red light.



\subsubsection{Results with or without Parameter Adjustment}
We further conduct ablation experiments to verify the performance of our parameter adjustment. We set three different sets of parameters for the vehicles to represent different user preferences for simulation details: agents have weak constraints on the velocity optimization energy term (G-1); agents have relatively balanced constraints on each optimization energy term (G-2); agents are more sensitive to collision avoidance (G-3). For each set of parameters, we simulate the same input data as in Section \ref{sec:res with SD} with or without our parameter adjustments, respectively. The difference between each set of simulation results and the ground truth in the distribution of velocity and steering angle are shown in Table \ref{table.PA}.

\begin{table}[b]
\centering  
\caption{Results with or without Parameter Adjustment on Different Parameters}  
\label{table.PA}
\setlength\tabcolsep{2.3pt} 
\begin{threeparttable}
\begin{tabular}{cccccccc}

\toprule[1pt] 
Group      & $w^{init}_{dir}$ & $w^{init}_{m}$ & $w^{init}_{e}$ & $w^{init}_{g}$ & $w^{init}_{c}$                  & Velocity & Steering Angle \\ \hline
G-1 w/o         & \multirow{2}{*}{1.0} & \multirow{2}{*}{0.9} & \multirow{2}{*}{1.3} & \multirow{2}{*}{1.0} & \multirow{2}{*}{1.0} & 0.5501    & 1.0864        \\
G-1 &                    &                    &                    &                    &                    & 0.5134   & 0.9973        \\ \hline
G-2 w/o         & \multirow{2}{*}{1.0} & \multirow{2}{*}{1.0} & \multirow{2}{*}{1.5} & \multirow{2}{*}{1.0} & \multirow{2}{*}{1.0} & 0.5329     & 1.0620        \\
G-2 &                    &                    &                    &                    &                    & 0.4647        & 0.9627             \\ \hline
G-3 w/o         & \multirow{2}{*}{0.8} & \multirow{2}{*}{1.1} & \multirow{2}{*}{1.4} & \multirow{2}{*}{1.0} & \multirow{2}{*}{1.3} & 0.6261    & 1.1810      \\
G-3 &                    &                    &                    &                    &                    & 0.5715   & 1.0215   \\
\bottomrule[1pt] 
\end{tabular}
\emph{Table 6 shows the differences of distribution between the simulation results and the ground truth under different parameters with or without parameter adjustment.}
\end{threeparttable}
\end{table}

By comparing the scores in Table \ref{table.PA}, it shows that the results with our parameter adjustment are closer to the ground truth distribution. During the simulation, our method dynamically changes the weight $w_{g}$ of the direction guidance term $E_{g}$ according to the agent's current position and destination, which enables the agent to move normally at irregular intersections without predefined lanes. In particular, when $w^{init}_{dir}$ is greater than $w^{init}_{g}$, some vehicles may not be able to steer without parameter adjustment. When agents interact with each other, we dynamically change the weights in the velocity term $E_{v}$ according to the characteristics of different types of agent interaction behavior, which makes the agent's behavior more realistic. Moreover, their trajectories are more flexible due to the different parameters of each agent.

The time cost of our parameter adjustment mainly comes from calculating the collision avoidance energy between interacting agents. We only use the agents' current velocities for computation, and the number of their neighbors is limited. In a crossroad scene with forty vehicles, the maximum computation time for our parameter adjustment is $0.088ms$, and the average computation time is $0.024ms$. Therefore, the time to compute the parameters has a negligible computational cost at each time step.

\subsubsection{Results with or without Planning-Aware Collision Avoidance}
\begin{figure}[t]
    \centering
    \includegraphics[height=3.85cm]{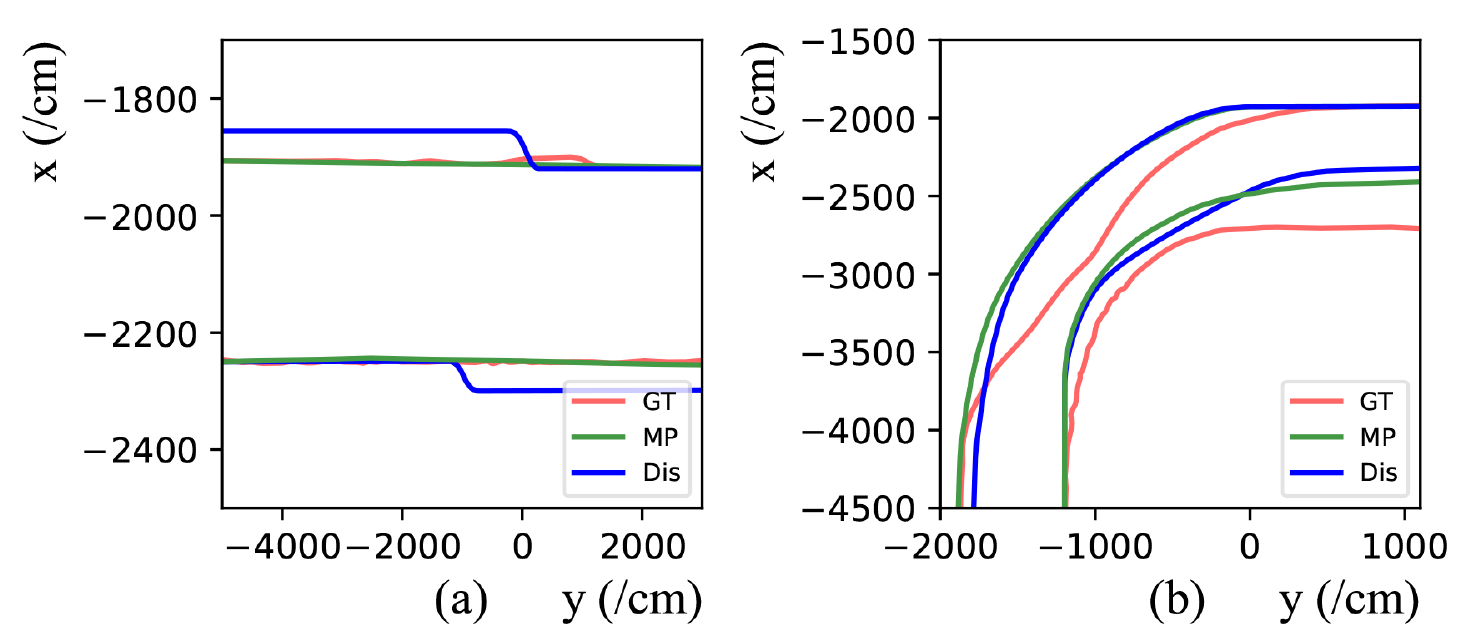}
    \caption{Trajectories of different collision avoidance strategies. Vehicles meet in the central area of the intersection:
    (a) straight vehicles from the opposite directions; (b) vehicles planned to turn left and right respectively.}
    \label{fig.mp}
\end{figure}

We intuitively compare the trajectories of agents to evaluate the performance of planning-aware collision avoidance. Among the vehicles that meet in the central area of the intersection, we select two circumstances that most intuitively reflect the impact of motion planning on collision avoidance: ($\rm \romannumeral 1$) straight vehicles from the opposite directions; ($\rm \romannumeral 2$) vehicles planned to turn left and right respectively. We use motion planning-aware collision avoidance and collision avoidance only based on safety distance  to simulate and draw their tracks respectively (Fig. \ref{fig.mp}).

Fig. \ref{fig.mp} (a) shows although the agents that only rely on distance to avoid collision have not collided, their trajectories have significantly deviated. As the relative distance between them is gradually decreasing, agents choose to increase the distance by turning to ensure that there is no collision. In Fig. \ref{fig.mp} (b), the trajectories of the two collision avoidance strategies are similar. However, the agents that only rely on distance to avoid collision have a temporary stagnation behavior. They choose to avoid getting closer to each other by slowing down to a standstill. These are all phenomena of over-safety. When ($\rm \romannumeral 1$) and ($\rm \romannumeral 2$) happen in the real world, drivers usually only increase their vigilance at the psychological level after considering the motion plannings of each other, and do not make additional avoidance behaviors.

\subsubsection{Results with or without Dataset Division} \label{sec: com datadiv}
We perform ablation experiments to verify the effectiveness of dataset division. The three simulation results are compared with statistical distribution of the ground truth in the crossroad scenario. The candidate velocity dataset is divided in the first simulation. All velocity data in the second and the third simulation is stored together, respectively. The difference is that the third simulation has a larger search range. Candidate velocities are sorted by the magnitude.

\begin{figure}[b]
    \centering
    \includegraphics[height=3.65cm]{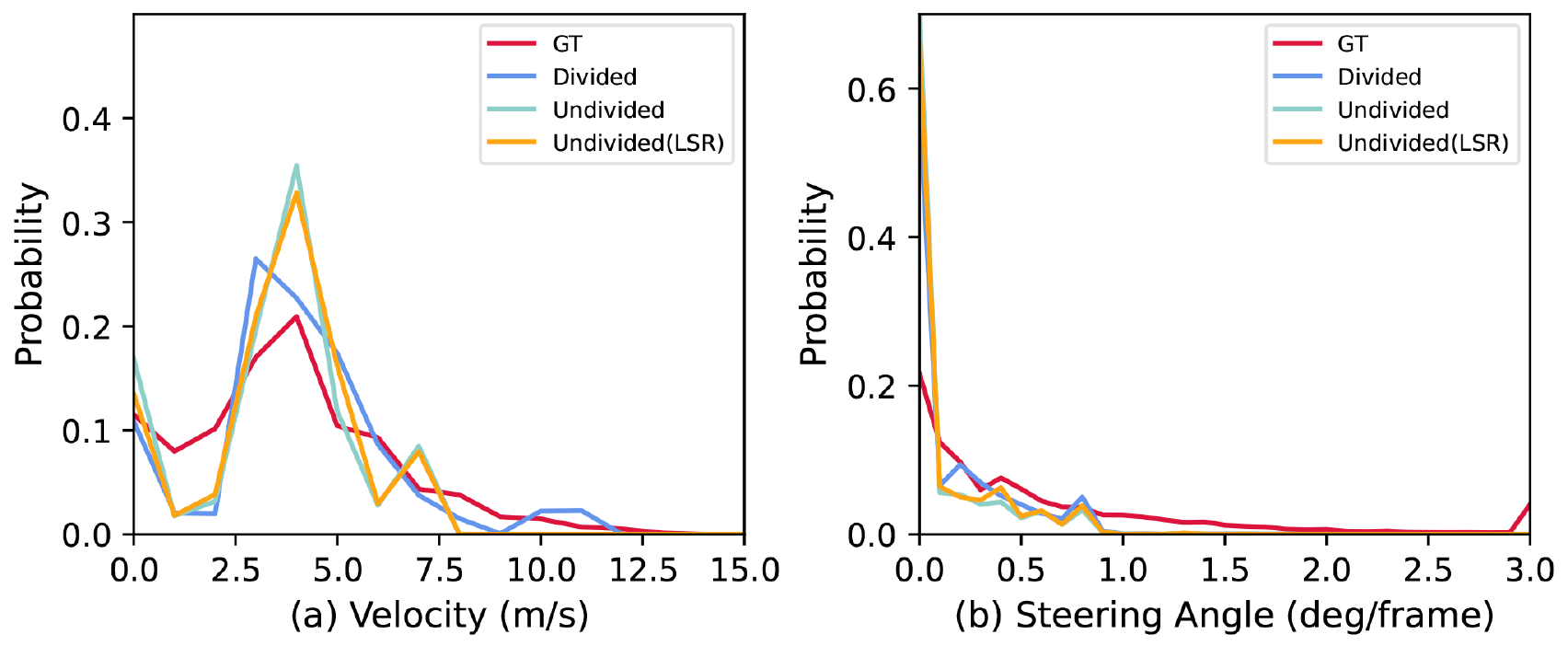}
    \caption{Results with or without dataset division. We compare simulation results with ground truth using dataset division, without dataset division, without dataset division but a larger search range, respectively.}
    \label{fig.exp_div}
\end{figure}

The experimental results are shown in Fig. \ref{fig.exp_div}.
Compared with the ground truth, the difference scores of velocity distribution of three simulation results are 0.4112: 0.5609: 0.5428, and the difference scores of steering angle distribution are 0.7593: 0.9714: 0.8974. Better simulation result is obtained after dataset division. As mentioned in Section \ref{sec:6.1}, we use the acceleration scheme proposed in \cite{ref9} to ensure the runtime performance of simulation. The search space of the agent is limited, whether or not dataset division is used. After the dataset division, there are more velocities in the search space of the agent at the current time that may be used to update its motion state. Although the performance can be improved by expanding the search range in the simulation without dataset division, evaluating too many candidate velocities at each timestep will lead to poor runtime performance.

\subsubsection{Comparisons with Other Methods} \label{sec: com baseline}
Previous methods mainly focus on freeways, which are difficult to be directly used to simulate intersection scenarios.
We compare our experimental results (in the crossroad scenario) with Heter-Sim \cite{ref9} and SUMO \cite{ref7}. The former is a state-of-the-art data-driven simulation method that can simulate heterogeneous multi-agent systems in different scenarios. Our method uses the same data-driven scheme as it. The latter is the famous model-driven microscopic traffic simulator, which supports users to customize traffic scenarios and simulate heterogeneous agents.

We randomly select two sets of trajectories with the same number from the crossroad scenario, including various types of agents from different directions. One set is randomly selected as the input data of our method and Heter-Sim\cite{ref9}. All methods are run twice to simulate two groups of ground truth. We initialize agents with the same number and locations as in the dataset. The desired speed is the average speed of the corresponding agent in the dataset. Our method uses the weights of the crossroad scene in Table \ref{tab. 2}, and we make some adjustments to the parameters in the Herter-Sim \cite{ref9} to fit our scenario.


\begin{figure}[t]
    \centering
    \includegraphics[height=6.55cm]{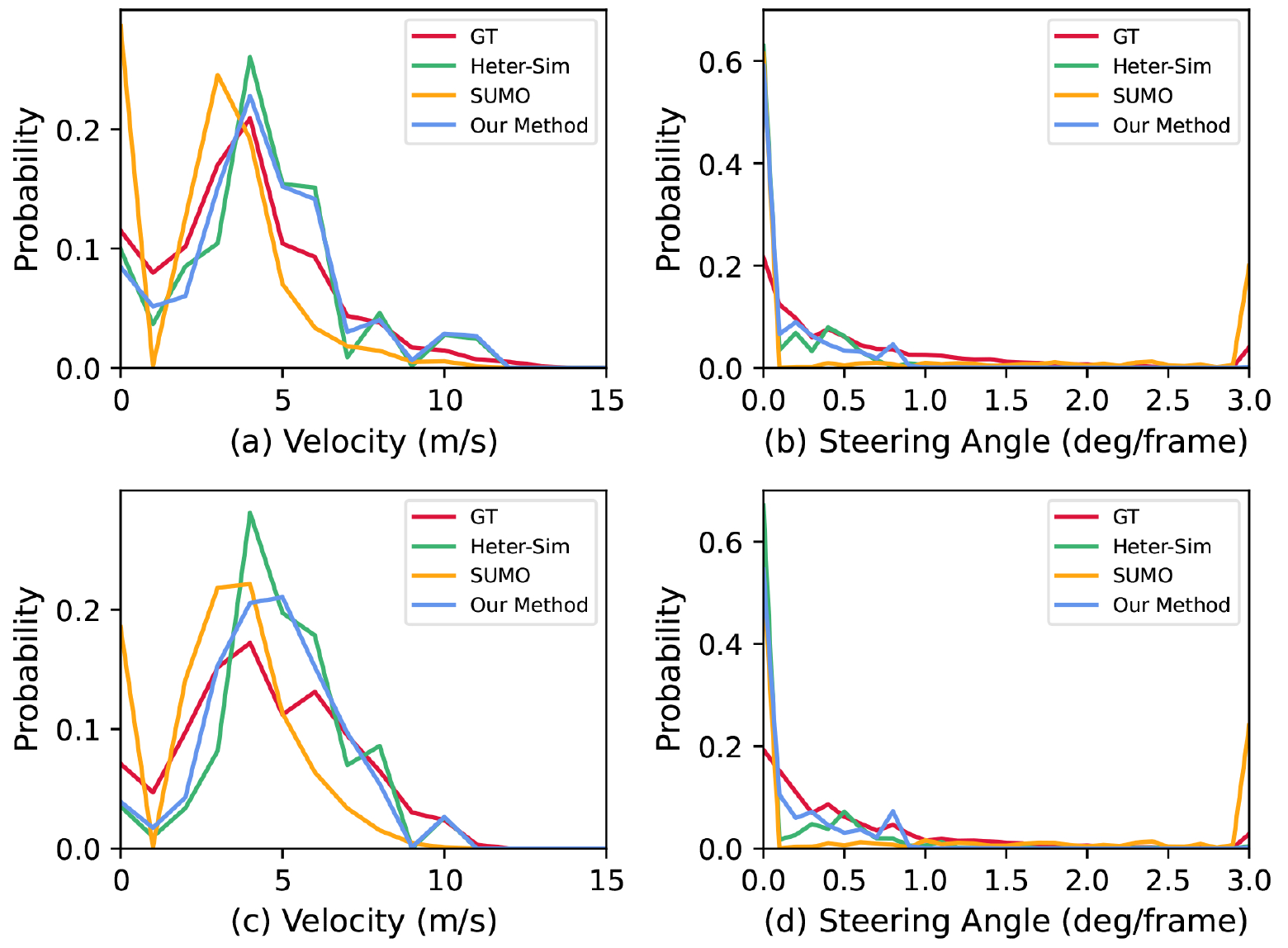}
    \caption{Comparisons with other methods. 
    Taking the same input, we compare the probability distributions of different ground truths simulated by our method and existing methods. (a)-(b): Simulated input data. (c)-(d): Simulated non-input data.}
    \label{fig.10}
\end{figure}

The various distributions for all methods are shown in Fig. \ref{fig.10}, and the detailed scores are shown in Table \ref{tab.4}. The velocity and steering angle distributions in our simulations are closer to the ground truth. 
In the above two simulation results, our method and SUMO \cite{ref7} have no obvious fluctuations. However, the scores of Heter-Sim \cite{ref9} changed significantly when simulating non-input data. Because it is a completely data-driven method, agents can only choose velocities from the input data to update their own motion state.
If the velocity of a vehicle in the control group is significantly different from the magnitude or direction of all the candidate velocities, the agent may not be able to simulate predetermined behavior. Due to the combination with velocity continuity model, our method can overcome this limitation.

In particular, both Heter-Sim \cite{ref9} and SUMO \cite{ref7} need predefined lanes. The former needs the assistance of the lane to guide the agent. In the latter, the agents pass in queues strictly along the lane line. Our method considers the motion planning in collision avoidance, and can dynamically adjust the the weight of terms in the energy function (Eq. \eqref{eq.8}) according to the surrounding during the simulation. Therefore, our method can plausibly simulate the movement of the agent in the intersection scene without predefined lanes in the central area.


\begin{table}[b]
\centering  
\caption{Simulation Results of Different Ground Truths}  
\label{tab.4}
\begin{threeparttable}
\setlength\tabcolsep{3.8pt} 
\begin{tabular}{ccccc}
\toprule[1pt] 
\multirow{2}{*}{} & \multicolumn{2}{c}{Input Data} & \multicolumn{2}{c}{Non-Input Data} \\ \cline{2-5} 
                  & Velocity    & Steering Angle   & Velocity      & Steering Angle     \\ \hline
Heter-Sim         &    0.3951         &      0.8368            &     0.5295          &  0.9749                  \\
SUMO       &    0.5414         &       1.1927           &    0.5508          &  1.2405                  \\
TraInterSim              & 0.3014            &    0.7754               &    0.3184           &        0.7630           
\\ 
\bottomrule[1pt] 
\end{tabular}
\emph{Table 7 shows the differences of velocity distribution and steering angle distribution between the simulation results and the real-world traffic data. Our method achieves lower scores, which demonstrates that the results are closer to real-world traffic than existing methods.}
\end{threeparttable}
\end{table}


\subsection{User Studies}

\begin{figure*}[!ht]
\begin{minipage}{0.49\linewidth}{}
\vspace{3pt}
\centerline{\includegraphics[totalheight=2.5cm]{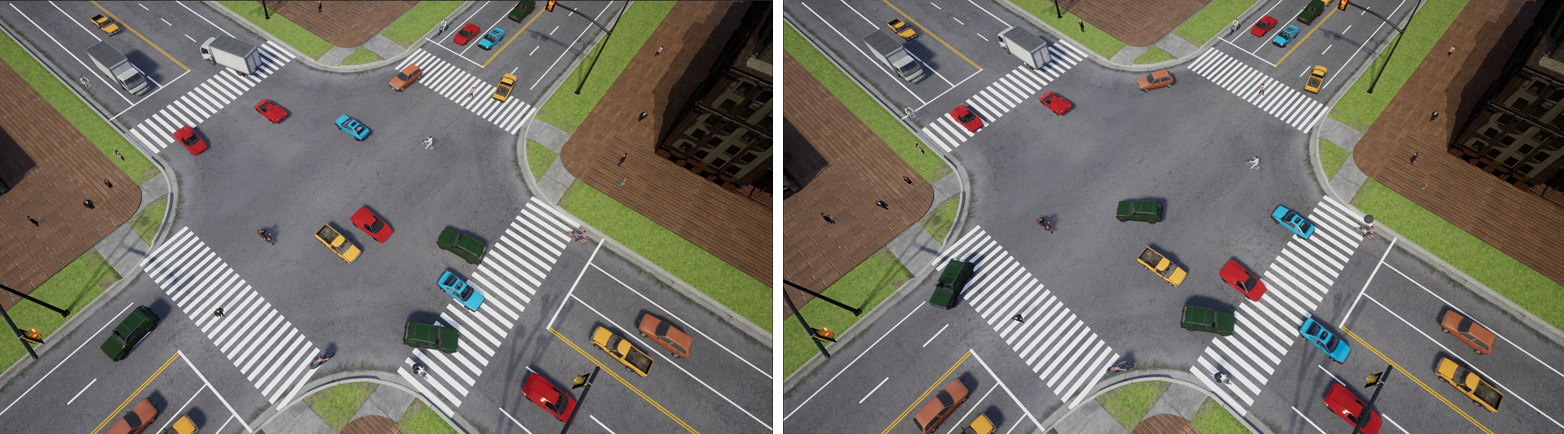}}
\centerline{(a)}
\end{minipage}
\begin{minipage}{0.505\linewidth}
\vspace{3pt}
\centerline{\includegraphics[totalheight=2.5cm]{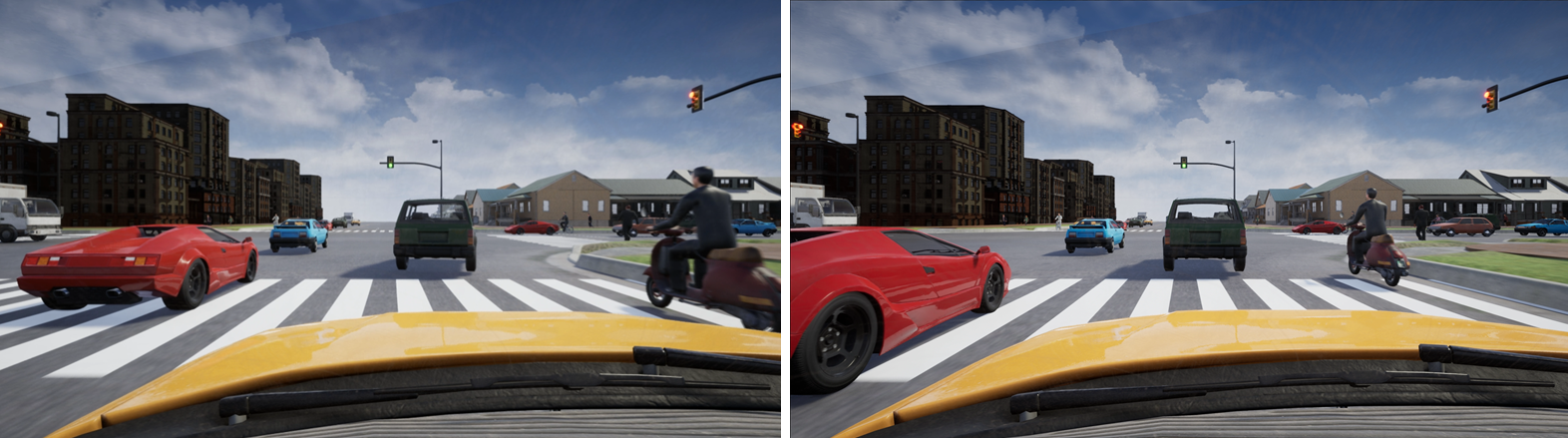}}
\centerline{(b)}
\end{minipage}

\caption{Snapshots of our street view study (a) and our agent view study (b). The left is the simulation result of Heter-Sim \cite{ref9} and the right is our simulation result. }
\label{fig.11}
\end{figure*}

To more intuitively evaluate our method, we design two paired comparison user studies. We generate traffic flows for different scenarios using our method and Heter-Sim \cite{ref9}, respectively. For each simulation we capture the traffic animation separately using street view (Fig. \ref{fig.11}a) and agent view (Fig. \ref{fig.11}b). The agent view can make the viewing experience more immersive for the user.

We use four intersectional scenarios for traffic simulations: 4-lane crossroad, 2-lane crossroad, T-junction, and Y-junction. We simulate these four traffic flows at two different crossroads (Crossroad-1, Crossroad-2), two of which are vehicles only, and the other two are mixed with heterogeneous agents. In the T-Junction scenario, all heterogeneous agents enter the intersection from different lanes. And in the Y-Junction scenario, only the vehicles are driving in the scene. 
For our method, the parameters in Table \ref{table1} are used for different scenarios. 
For Heter-Sim \cite{ref9}, the parameters used in different scenarios are listed in Table \ref{tab.5}. 
The input data used by the two methods in different scenarios is consistent with Section \ref{sec: com baseline}. That is, we generate 24 traffic intersection animations and 12 comparison pairs for this user study.

\begin{figure*}[t]
\begin{minipage}{0.49\linewidth}{}
\vspace{3pt}
\centerline{\includegraphics[totalheight=4.5cm]{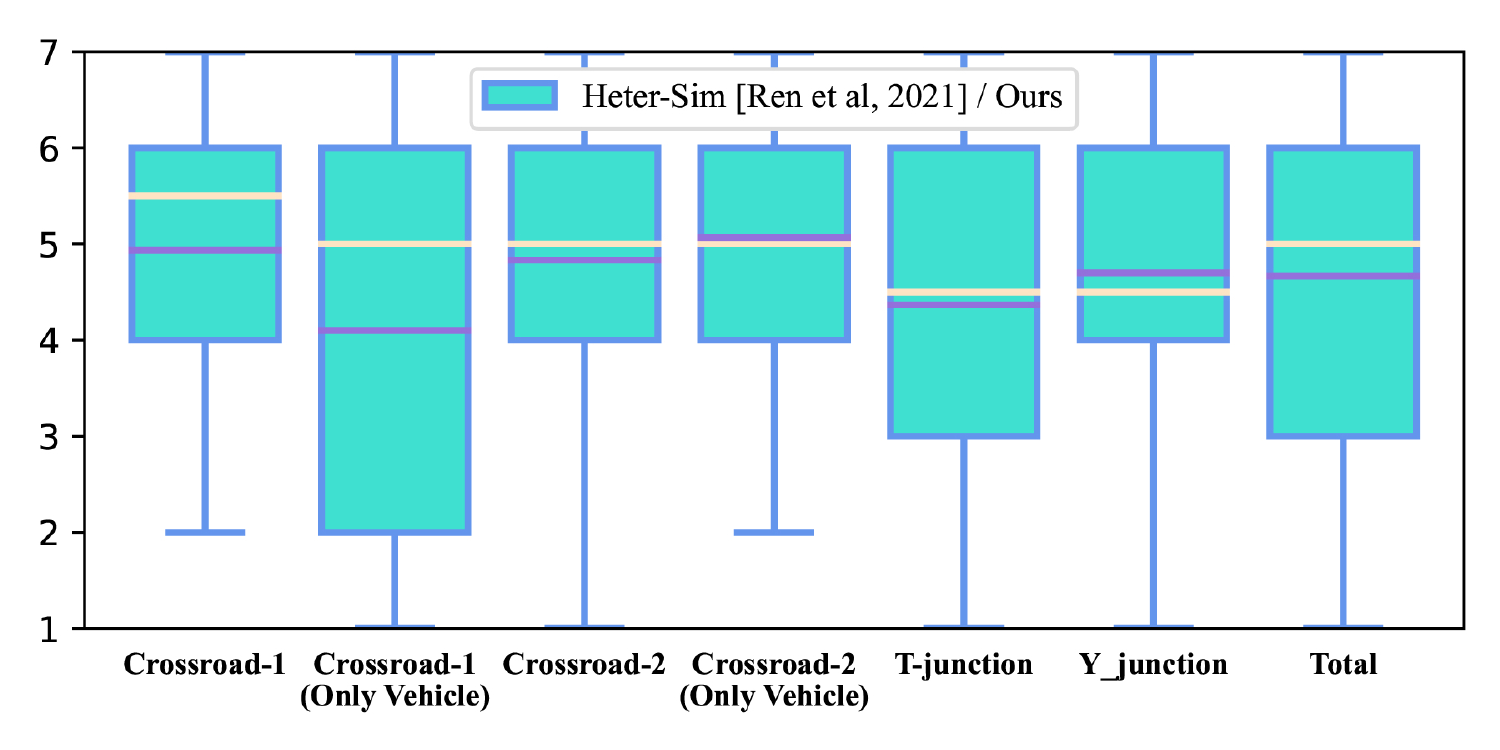}}
\centerline{(a)}
\end{minipage}
\begin{minipage}{0.505\linewidth}
\vspace{3pt}
\centerline{\includegraphics[totalheight=4.5cm]{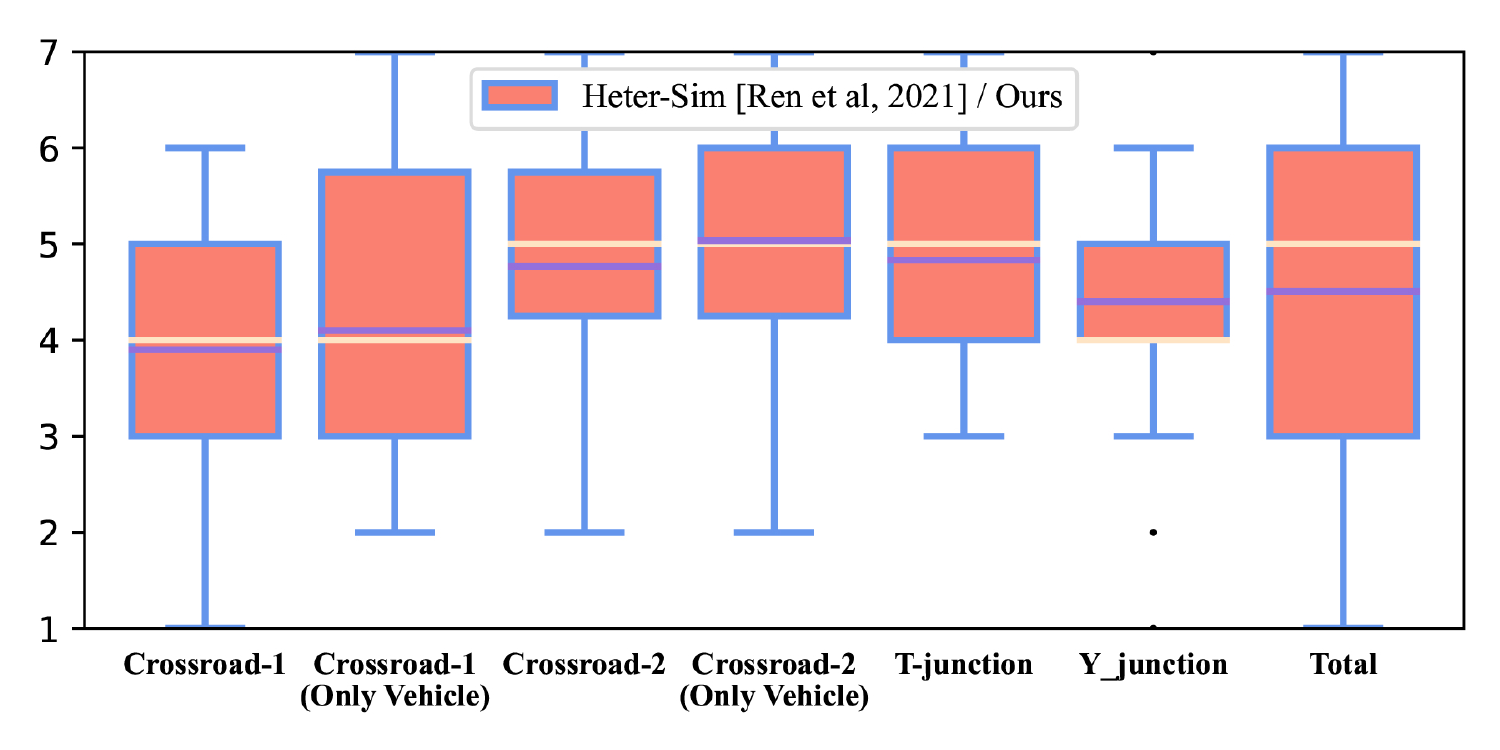}}
\centerline{(b)}
\end{minipage}
\caption{Scores of our user study. The 7-point Likert scale is introduced into our user study to measure users' preferences for the plausibility of the simulation results of different methods. The lower the score, the more participants prefer the method on the left; the higher the score indicates that participants prefer the method on the right. (a) Detailed scores for the street view study. (b) Detailed scores for the agent view study.}
    \label{fig.12}
\end{figure*}

Thirty participants are recruited for our user study. All of them are graduate students from the same university, including eighteen males and twelve females. In each comparative group of user studies, we use the 7-point Likert scale to ask participants to rate the simulation results from 1 to 7. The score indicates a preference for the plausibility of the results, 1 indicates that the results on the left are completely plausible, 7 indicates that the results on the right are completely plausible, and 4 indicates that there is no preference for the results on both sides. Fig. \ref{fig.12} shows the detailed scores of each comparison pair.

\emph{The street view study:} The detailed scores of thirty participants for the street view study are shown in Fig. \ref{fig.12}a. In this figure, in addition to the score of each scene, we give the total voting results. It can be seen from the figure that participants generally have higher evaluation of our method. Further, we perform the one-sample t-test to quantify the statistical significance of the voting results. We assume that the mean score of our method is greater than 4 (no preference). The experimental results show that the mean scores of the plausibility of our method in the street view study are significantly higher than that of the comparison method, which is statistically significant ($t(29) = 2.2534$, $p = 0.0159 < 0.05$).

\renewcommand{\arraystretch}{1.4}
\begin{table}[b]
\centering  
\caption{The Parameters of Heter-Sim in the User Study }  
\label{tab.5}
\begin{threeparttable}
\setlength\tabcolsep{2.7pt} 

\begin{tabular}{cccccccccc}
\toprule[1pt] 

\multicolumn{2}{c}{Scenario}                                 & $E{\rm^{dir}_{t}}$ & $E{\rm^{L}_{t}}$ & $E{\rm^{Ins}_{c}}$ & $E{\rm^{Anti}_{c}}$ & $E{\rm_{a}}$ &  $E{\rm_{d}}$ &  $E{\rm_{p}}$ &  $E{\rm_{sc}}$\\ \hline
\multirow{3}{*}{Crossroad-1}                      & Car        & 1.0  & 1.0  & 1.0  & 0.5  & 0.0  & 5.0  & 5.0  & 2.0\\
                                                & Pedestrian & 0.5    & 1.0   & 1.0 & 1.0  & 0.5  & 1.5  & 2.0  & 1.5  \\
                                                & Bicycle    & 3.0  & 1.0  & 1.0  & 1.0  & 0.0 & 5.0 & 10.0  & 2.0  \\ 
                                                \hline
\multirow{3}{*}{Crossroad-2}                      & Car        & 1.0  & 1.0  & 1.0  & 0.5  & 0.0  & 5.0  & 5.0  & 5.0\\
                                                & Pedestrian & 0.5  & 1.0  & 1.0  & 1.0  & 1.0 &2.0 & 3.0 &1.5  \\
                                                & Bicycle    & 2.0  & 1.0  & 1.0  & 1.0  & 0.0 & 6.0  & 10.0  & 2.0  \\\hline
\multicolumn{1}{l}{\multirow{3}{*}{T-junction}} & Car        & 1.0  & 1.0  & 1.0  & 0.5  & 0.0   & 5.0  & 10.0  & 2.0 \\
\multicolumn{1}{l}{}                            & Pedestrian & 0.5  & 1.0  & 1.0  & 1.0  & 1.0 & 1.5 & 3.0 &1.5 \\
\multicolumn{1}{l}{}                            & Bicycle    & 1.5  & 1.0  & 1.0  & 1.0  & 0.0  & 3.0  & 10.0  & 2.0\\ \hline
Y-junction                                      & Car        & 1.0  & 1.0  & 1.0  & 0.5 & 0.0  & 5.0  & 5.0  & 5.0 \\

\bottomrule[1pt]

\end{tabular}
\emph{Table 8 presents the parameters of Heter-Sim in our user study in different scenarios. }
\end{threeparttable}
\end{table}

\emph{The agent view study:} The detailed scores of the thirty participants of the agent view study are shown in Fig. \ref{fig.12}b, in which we also present the total voting results in addition to the scores for each scene. Our method also achieves better results, and the difference between our method and the comparison method is more pronounced than that of the street view study. The result of the one-sample t-test indicate that the mean scores of our method differ significantly from the hypothesized mean ($t(29) = 1.8454$, $p = 0.0376 < 0.05$).

Our method achieves relatively better evaluations in both of the above user studies. After the user study, we conducted an informal interview with the participants. The participants indicate that the reason why they think our method is better mainly is that our method has smoother trajectories and interactions. However, since our motion planning-aware collision avoidance scheme reduces the sensitivity of agents to some their neighbors. Participants think that although our method has no collision in interaction, sometimes the distance between agents is slightly closer.

\section{Conclusion}
We present a novel adaptive and planning-aware hybrid-driven simulation method for traffic intersection scenarios. We show our simulation results in different intersection scenarios. To validate our approach, we conduct comparison experiments and user studies. Experimental outcomes show that our method achieves more realistic results than the state-of-the-art data-driven traffic simulation method.

The quantity and quality of datasets has always been an important factor limiting the performance of data-driven methods. Our hybrid-driven method can generate plausible simulation results with flawed input samples, and velocities that do not exist in the dataset will be supplemented. Our trajectory optimization method can be used to eliminate the noise in the input data, and it can also be extended to other works to improve the quality of the sampled data. Our method can ensure that agents obey microscopic traffic rules (e.g., safe gaps, traffic lights) and maintain movement characteristics consistent with the real world in the intersection area where lanes are not clearly defined. Instead of only considering the inter-agent relative distance, we inject the motion planning into the collision avoidance energy function, which is more in line with the logic of real-world drivers to make interactive decisions. Furthermore, we adaptively change the parameters during the simulation with a set of benchmark parameters. This ensures that homogeneous agents have the same movement patterns without losing individual flexibility of the agent.

There are still some limitations of our approach. Our data supplement scheme is based on the continuity of the agent's velocity at the previous time step, which is sometimes imprecise. Another limitation of our method is that it is unable to simulate large-scale traffic scenes at an interactive rate. As we analyzed in Section \ref{sec:6.1}, when a large number of agents enter the intersection area, we need to spend a lot of time computing the interactions between agents. In the future, we plan to introduce parallel implementation and GPU-acceleration to speed up our method.

There are still many ways to extend our method in the future. We plan to extend our method to be editable, which automatically generate a road network based on the topology designed by the user. Algorithmically, in addition to overcome our existing limitations, deep learning can be used to more accurately model the interactions between agents. We also consider adding psychological characteristics or other sensory information such as vision \cite{ref5} and hearing \cite{ref52} can also be introduced.


%



\ifCLASSOPTIONcaptionsoff
  \newpage
\fi

\bibliographystyle{IEEEtran}
\bibliography{ref}

\begin{thebibliography}{10}
\providecommand{\url}[1]{#1}
\csname url@samestyle\endcsname
\providecommand{\newblock}{\relax}
\providecommand{\bibinfo}[2]{#2}
\providecommand{\BIBentrySTDinterwordspacing}{\spaceskip=0pt\relax}
\providecommand{\BIBentryALTinterwordstretchfactor}{4}
\providecommand{\BIBentryALTinterwordspacing}{\spaceskip=\fontdimen2\font plus
\BIBentryALTinterwordstretchfactor\fontdimen3\font minus
  \fontdimen4\font\relax}
\providecommand{\BIBforeignlanguage}[2]{{%
\expandafter\ifx\csname l@#1\endcsname\relax
\typeout{** WARNING: IEEEtran.bst: No hyphenation pattern has been}%
\typeout{** loaded for the language `#1'. Using the pattern for}%
\typeout{** the default language instead.}%
\else
\language=\csname l@#1\endcsname
\fi
#2}}
\providecommand{\BIBdecl}{\relax}
\BIBdecl

\bibitem{ref8}
Q.~Chao, Z.~Deng, J.~Ren, Q.~Ye, and X.~Jin, ``Realistic data-driven traffic
  flow animation using texture synthesis,'' \emph{{IEEE} Trans. Vis. Comput.
  Graph.}, vol.~24, no.~2, pp. 1167--1178, 2018.

\bibitem{ref10}
J.~Sewall, J.~P. van~den Berg, M.~C. Lin, and D.~Manocha, ``Virtualized
  traffic: Reconstructing traffic flows from discrete spatiotemporal data,''
  \emph{{IEEE} Trans. Vis. Comput. Graph.}, vol.~17, no.~1, pp. 26--37, 2011.

\bibitem{ref24}
J.~Shen and X.~Jin, ``Detailed traffic animation for urban road networks,''
  \emph{Graph. Model.}, vol.~74, no.~5, pp. 265--282, 2012.

\bibitem{ref53}
W.~Li, D.~Wolinski, and M.~C. Lin, ``City-scale traffic animation using
  statistical learning and metamodel-based optimization,'' \emph{{ACM} Trans.
  Graph.}, vol.~36, no.~6, pp. 200:1--200:12, 2017.

\bibitem{ref1}
Q.~Chao, X.~Jin, H.~Huang, S.~Foong, L.~Yu, and S.~Yeung, ``Force-based
  heterogeneous traffic simulation for autonomous vehicle testing,'' in
  \emph{Proc IEEE Int Conf Rob Autom}, 2019, pp. 8298--8304.

\bibitem{ref2}
I.~Karamouzas, B.~Skinner, and S.~J. Guy, ``Universal power law governing
  pedestrian interactions,'' \emph{Phys. Rev. Lett.}, vol. 113, no.~23, p.
  238701, 2014.

\bibitem{ref3}
J.~P. van~den Berg, M.~C. Lin, and D.~Manocha, ``Reciprocal velocity obstacles
  for real-time multi-agent navigation,'' in \emph{Proc IEEE Int Conf Rob
  Autom}, 2008, pp. 1928--1935.

\bibitem{ref4}
J.~van~den Berg, S.~J. Guy, M.~C. Lin, and D.~Manocha, ``Reciprocal
  \emph{n}-body collision avoidance,'' in \emph{Proc. 14th Int. Symp. Robotics
  Research}, 2011, pp. 3--19.

\bibitem{ref5}
J.~Ondrej, J.~Pettr{\'{e}}, A.~Olivier, and S.~Donikian, ``A synthetic-vision
  based steering approach for crowd simulation,'' \emph{{ACM} Trans. Graph.},
  vol.~29, no.~4, pp. 123:1--123:9, 2010.

\bibitem{ref6}
P.~A. Lopez, M.~Behrisch, L.~Bieker-Walz, J.~Erdmann, Y.-P. Fl{\"o}tter{\"o}d,
  R.~Hilbrich, L.~L{\"u}cken, J.~Rummel, P.~Wagner, and E.~Wie{\ss}ner,
  ``Microscopic traffic simulation using sumo,'' \emph{IEEE Intelligent
  Transportation Systems Conference (ITSC)}, 2018.

\bibitem{ref7}
\BIBentryALTinterwordspacing
``Mit intelligent transportation systems,'' 2011. [Online]. Available:
  \url{its.mit.edu/}
\BIBentrySTDinterwordspacing

\bibitem{ref11}
D.~Wilkie, J.~Sewall, and M.~C. Lin, ``Flow reconstruction for data-driven
  traffic animation,'' \emph{{ACM} Trans. Graph.}, vol.~32, no.~4, pp.
  89:1--89:10, 2013.

\bibitem{ref12}
W.~Li, D.~Nie, D.~Wilkie, and M.~C. Lin, ``Citywide estimation of traffic
  dynamics via sparse {GPS} traces,'' \emph{{IEEE} Intell. Transp. Syst. Mag.},
  vol.~9, no.~3, pp. 100--113, 2017.

\bibitem{ref13}
W.~Li, M.~Jiang, Y.~Chen, and M.~C. Lin, ``Estimating urban traffic states
  using iterative refinement and wardrop equilibria,'' \emph{IET Intel.
  Transport Syst.}, vol.~12, no.~8, pp. 875--883, 2018.

\bibitem{ref55}
A.~Sarkar, K.~Czarnecki, M.~Angus, C.~Li, and S.~Waslander, ``Trajectory
  prediction of traffic agents at urban intersections through learned
  interactions,'' pp. 1--8, 2017.

\bibitem{ref56}
X.~Huang, S.~G. McGill, B.~C. Williams, L.~Fletcher, and G.~Rosman,
  ``Uncertainty-aware driver trajectory prediction at urban intersections,''
  pp. 9718--9724, 2019.

\bibitem{ref57}
W.~Li, D.~Wolinski, and M.~C. Lin, ``Adaps: Autonomous driving via principled
  simulations,'' pp. 7625--7631, 2019.

\bibitem{ref58}
L.~Lin, W.~Li, H.~Bi, and L.~Qin, ``Vehicle trajectory prediction using lstms
  with spatial–temporal attention mechanisms,'' \emph{IEEE Intelligent
  Transportation Systems Magazine}, vol.~14, no.~2, pp. 197--208, 2022.

\bibitem{ref14}
G.~Berseth, P.~Faloutsos, P.~Faloutsos, and P.~Faloutsos, ``Steerfit: Automated
  parameter fitting for steering algorithms,'' in \emph{Proc. ACM
  SIGGRAPH/Eurographics Symp. Comput. Animation}, 2014, pp. 113--122.

\bibitem{ref15}
D.~Wolinski, S.~J. Guy, A.~Olivier, M.~C. Lin, D.~Manocha, and J.~Pettr{\'{e}},
  ``Parameter estimation and comparative evaluation of crowd simulations,''
  \emph{Comput. Graph. Forum}, vol.~33, no.~2, pp. 303--312, 2014.

\bibitem{ref16}
Q.~Chao, P.~Liu, Y.~Han, Y.~Lin, C.~Li, Q.~Miao, and X.~Jin, ``A calibrated
  force-based model for mixed traffic simulation,'' \emph{{IEEE} Trans. Vis.
  Comput. Graph.}, pp. 1--1, 2021.

\bibitem{ref17}
A.~Kesting and M.~Treiber, ``Calibrating car-following models using trajectory
  data: Methodological study,'' \emph{Transp. Res. Rec.: J. Transp. Res.
  Board}, vol. 2088, no.~1, pp. 148--156, 2008.

\bibitem{ref9}
J.~Ren, W.~Xiang, Y.~Xiao, R.~Yang, D.~Manocha, and X.~Jin, ``Heter-sim:
  Heterogeneous multi-agent systems simulation by interactive data-driven
  optimization,'' \emph{{IEEE} Trans. Vis. Comput. Graph.}, vol.~27, no.~3, pp.
  1953--1966, 2021.

\bibitem{ref18}
Q.~Chao, Z.~Deng, and X.~Jin, ``Vehicle-pedestrian interaction for mixed
  traffic simulation,'' \emph{Comput. Animat. Virtual Worlds}, vol.~26, no.
  3-4, pp. 405--412, 2015.

\bibitem{ref19}
G.~F. Newell, ``Nonlinear effects in the dynamics of car following,''
  \emph{Operations Research}, vol.~9, no.~2, pp. 209--229, 1961.

\bibitem{ref20}
K.~Nagel and M.~Schreckenberg, ``A cellular automaton model for freeway
  traffic,'' \emph{J. Physics I}, vol.~2, no.~12, pp. 2221--2229, 1992.

\bibitem{ref21}
L.~A. Pipes, ``An operational analysis of traffic dynamics,'' \emph{J. Appl.
  Phys.}, vol.~24, no.~3, pp. 274--281, 1953.

\bibitem{ref22}
M.~Treiber and D.~Helbing, ``Microsimulations of freeway traffic including
  control measures,'' \emph{Automatisierungstechnik}, vol.~49, no.~11, 2002.

\bibitem{ref23}
A.~Kesting, M.~Treiber, and D.~Helbing, ``General lane-changing model mobil for
  car-following models,'' \emph{Transp. Res. Rec.: J. Transp. Res. Board}, vol.
  1999, pp. 86--94, 2007.

\bibitem{ref25}
X.~Yang, W.~Su, J.~Deng, X.~Jin, G.~Tan, and Z.~Pan, ``Real-virtual fusion
  model for traffic animation,'' \emph{Comput. Animat. Virtual Worlds},
  vol.~28, no.~6, 2017.

\bibitem{ref26}
T.~Mao, H.~Wang, Z.~Deng, and Z.~Wang, ``An efficient lane model for complex
  traffic simulation,'' \emph{Comput. Animat. Virtual Worlds}, vol.~26, no.
  3-4, pp. 397--403, 2015.

\bibitem{ref27}
J.~Sewall, D.~Wilkie, P.~Merrell, and M.~C. Lin, ``Continuum traffic
  simulation,'' \emph{Comput. Graph. Forum}, vol.~29, no.~2, pp. 439--448,
  2010.

\bibitem{ref28}
V.~Shvetsov and D.~Helbing, ``Macroscopic dynamics of multilane traffic,''
  \emph{Phys. Rev. E}, vol.~59, no.~6, pp. 6328--6339, 1999.

\bibitem{ref29}
M.~J. Lighthill and G.~B. Whitham, ``On kinematic waves. ii. a theory of
  traffic flow on long crowded roads,'' \emph{Proc. R. Soc. A Math. Phys. Eng.
  Sci.}, vol. 229, no. 1178, pp. 317--345, 1955.

\bibitem{ref30}
H.~M. Zhang, ``A non-equilibrium traffic model devoid of gas-like behavior,''
  \emph{Transp. Res. Part B Methodol.}, vol.~36, no.~3, pp. 275--290, 2002.

\bibitem{ref31}
J.~Sewall, D.~Wilkie, and M.~C. Lin, ``Interactive hybrid simulation of
  large-scale traffic,'' \emph{{ACM} Trans. Graph.}, vol.~30, no.~6, p. 135,
  2011.

\bibitem{ref36}
Q.~Chao, J.~Shen, and X.~Jin, ``Video-based personalized traffic learning,''
  \emph{Graph. Model.}, vol.~75, no.~6, pp. 305--317, 2013.

\bibitem{ref38}
H.~Bi, T.~Mao, Z.~Wang, and Z.~Deng, ``A data-driven model for lane-changing in
  traffic simulation,'' in \emph{Proc. ACM SIGGRAPH/ Eurograph. Symp. Comput.
  Animation}, 2016, pp. 149--158.

\bibitem{ref50}
Y.~Zhang, W.~Wang, W.~Guo, P.~Lv, M.~Xu, W.~Chen, and D.~Manocha, ``D2-tpred:
  Discontinuous dependency for trajectory prediction under traffic lights,'' in
  \emph{Proc. Eur. Conf. Comput. Vis.}, vol. 13668, 2022, pp. 522--539.

\bibitem{ref37}
H.~Bi, T.~Mao, Z.~Wang, and Z.~Deng, ``A deep learning-based framework for
  intersectional traffic simulation and editing,'' \emph{{IEEE} Trans. Vis.
  Comput. Graph.}, vol.~26, no.~7, pp. 2335--2348, 2020.

\bibitem{ref39}
W.~van Toll and J.~Pettr{\'{e}}, ``Algorithms for microscopic crowd simulation:
  Advancements in the 2010s,'' \emph{Comput. Graph. Forum}, vol.~40, no.~2, pp.
  731--754, 2021.

\bibitem{ref40}
D.~Helbing and P.~Molnar, ``Social force model for pedestrian dynamics,''
  \emph{Phys.Rev.E}, vol.~51, no.~5, p. 4282, 1995.

\bibitem{ref41}
D.~Helbing, I.~Farkas, and T.~Vicsek, ``Simulating dynamical features of escape
  panic,'' \emph{Nature}, vol. 407, no. 6803, p. 2000, 2000.

\bibitem{ref42}
Y.~Han, Q.~Chao, and X.~Jin, ``A simplified force model for mixed traffic
  simulation,'' \emph{Comput. Animat. Virtual Worlds}, vol.~32, no.~1, 2021.

\bibitem{ref43}
I.~Karamouzas, N.~Sohre, R.~Narain, and S.~J. Guy, ``Implicit crowds:
  optimization integrator for robust crowd simulation,'' \emph{{ACM} Trans.
  Graph.}, vol.~36, no.~4, pp. 136:1--136:13, 2017.

\bibitem{ref44}
F.~Durupinar, U.~G{\"{u}}d{\"{u}}kbay, A.~Aman, and N.~I. Badler,
  ``Psychological parameters for crowd simulation: From audiences to mobs,''
  \emph{{IEEE} Trans. Vis. Comput. Graph.}, vol.~22, no.~9, pp. 2145--2159,
  2016.

\bibitem{ref45}
J.~H. Park, F.~A. Rojas, and H.~S. Yang, ``A collision avoidance behavior model
  for crowd simulation based on psychological findings,'' \emph{Comput. Animat.
  Virtual Worlds}, vol.~24, no. 3-4, pp. 173--183, 2013.

\bibitem{ref46}
S.~J. Guy, S.~Kim, M.~C. Lin, and D.~Manocha, ``Simulating heterogeneous crowd
  behaviors using personality trait theory,'' in \emph{Proc. - SCA: ACM
  SIGGRAPH / Eurographics Symp. Comput. Anim.}, 2011.

\bibitem{ref48}
B.~Zhou, X.~Wang, and X.~Tang, ``Understanding collective crowd behaviors:
  Learning a mixture model of dynamic pedestrian-agents,'' in \emph{Proc IEEE
  Comput Soc Conf Comput Vision Pattern Recognit}, 2012, pp. 2871--2878.

\bibitem{ref49}
D.~Yang, X.~Zhou, G.~Su, and S.~Liu, ``Model and simulation of the
  heterogeneous traffic flow of the urban signalized intersection with an
  island work zone,'' \emph{{IEEE} Trans. Intell. Transp. Syst.}, vol.~20,
  no.~5, pp. 1719--1727, 2019.

\bibitem{ref51}
S.~Ettinger, S.~Cheng, B.~Caine, C.~Liu, H.~Zhao, S.~Pradhan, Y.~Chai, B.~Sapp,
  C.~R. Qi, Y.~Zhou, Z.~Yang, A.~Chouard, P.~Sun, J.~Ngiam, V.~Vasudevan,
  A.~McCauley, J.~Shlens, and D.~Anguelov, ``Large scale interactive motion
  forecasting for autonomous driving : The waymo open motion dataset,'' in
  \emph{Proc IEEE Int Conf Comput Vision}, 2021, pp. 9690--9699.

\bibitem{ref54}
Q.~Chao, H.~Bi, W.~Li, T.~Mao, Z.~Wang, M.~C. Lin, and Z.~Deng, ``A survey on
  visual traffic simulation: Models, evaluations, and applications in
  autonomous driving,'' \emph{Comput. Graph. Forum}, vol.~39, no.~1, pp.
  287--308, 2020.

\bibitem{ref52}
Y.~Wang, M.~Kapadia, P.~Huang, L.~Kavan, and N.~I. Badler, ``Sound localization
  and multi-modal steering for autonomous virtual agents,'' in \emph{Proc Symp
  Interactive 3D Graphics}, 2014, pp. 23--30.

\end{thebibliography}

$\quad$

$\quad$

$\quad$

$\quad$

$\quad$

$\quad$

$\quad$

$\quad$

$\quad$

$\quad$

$\quad$

$\quad$

$\quad$

$\quad$

$\quad$

\begin{IEEEbiography}[{\includegraphics[width=1in,height=1.25in,clip,keepaspectratio]{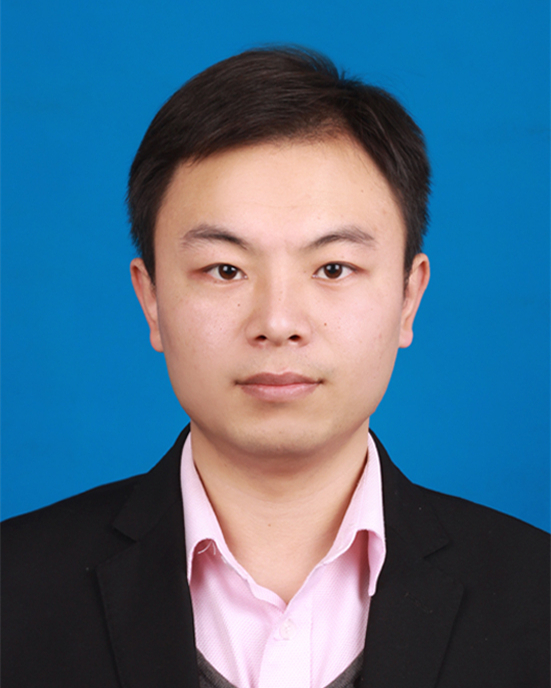}}]
{Pei Lv} received the Ph.D. degree from the State Key Laboratory of CAD\&CG, Zhejiang University, Hangzhou, China, in 2013. He is an Associate Professor with the School of Computer and Artificial Intelligence, Zhengzhou University, Zhengzhou, China. His research interests include computer vision and computer graphics. He has authored more than 50 journal and conference papers in the above areas, including the IEEE TRANSACTIONS ON IMAGE PROCESSING , the IEEE TRANSACTIONS ON CIRCUITS AND SYSTEMS FOR VIDEO TECHNOLOGY, the IEEE TRANSACTIONS ON AFFECTIVE COMPUTING, CVPR, ECCV, ACM MM, et al..
\end{IEEEbiography}

\begin{IEEEbiography}[{\includegraphics[width=1in,height=1.4in,clip,keepaspectratio]{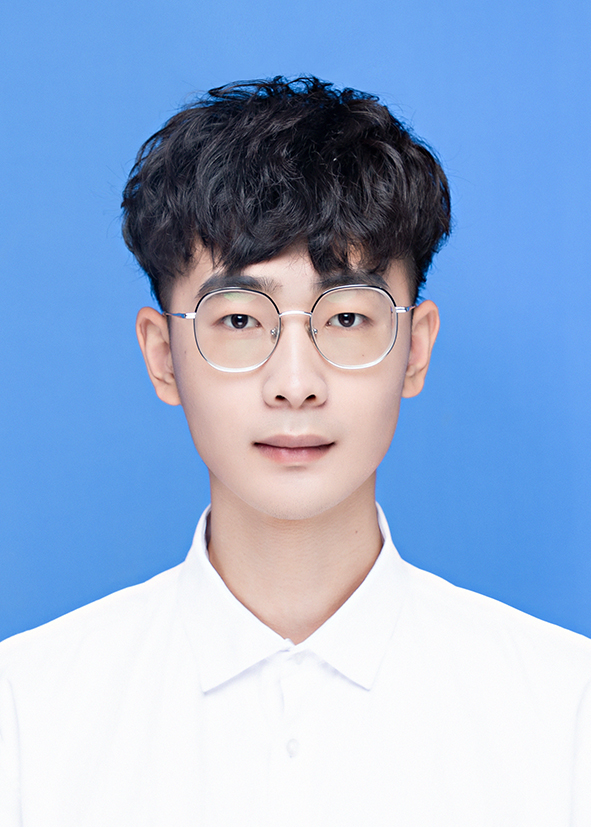}}]
{Xinming Pei} received the BSc. degree in the network engineering, Henan University, China, in 2019. He is working toward the MSc. degree at the School of Computer and Artificial Intelligence, Zhengzhou University, Zhengzhou, China. His research interests include physics-based simulation and crowd simulation.
\end{IEEEbiography}

\begin{IEEEbiography}[{\includegraphics[width=1in,height=1.25in,clip,keepaspectratio]{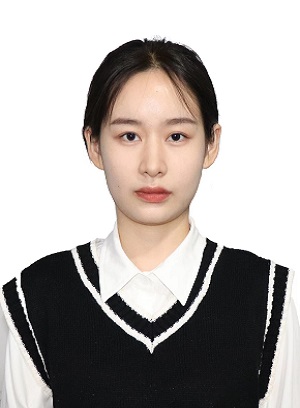}}]
{Xinyu Ren} received the BSc. degree in software engineering, Zhengzhou University, China, in 2021. She is working toward the MSc. degree at the School of Computer and Artificial Intelligence, Zhengzhou University, Zhengzhou, China. Her main research interest is traffic simulation.
\end{IEEEbiography}

\begin{IEEEbiography}[{\includegraphics[width=1in,height=1.25in,clip,keepaspectratio]{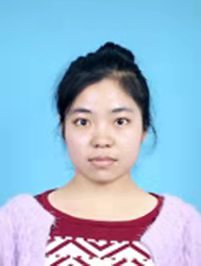}}]
{Yuzhen Zhang} received the BSc. and MSc. degrees in software engineering from Henan Polytechnic University, Jiaozuo, China. She is currently pursuing the Ph.D. degree in School of Computer and Artificial Intelligence, Zhengzhou University, Zhengzhou, China. Her current research interests include machine learning, computer vision and their applications to motion prediction, scene understanding, and interaction modeling for intelligent autonomous systems.
\end{IEEEbiography}

\begin{IEEEbiography}[{\includegraphics[width=1in,height=1.25in,clip,keepaspectratio]{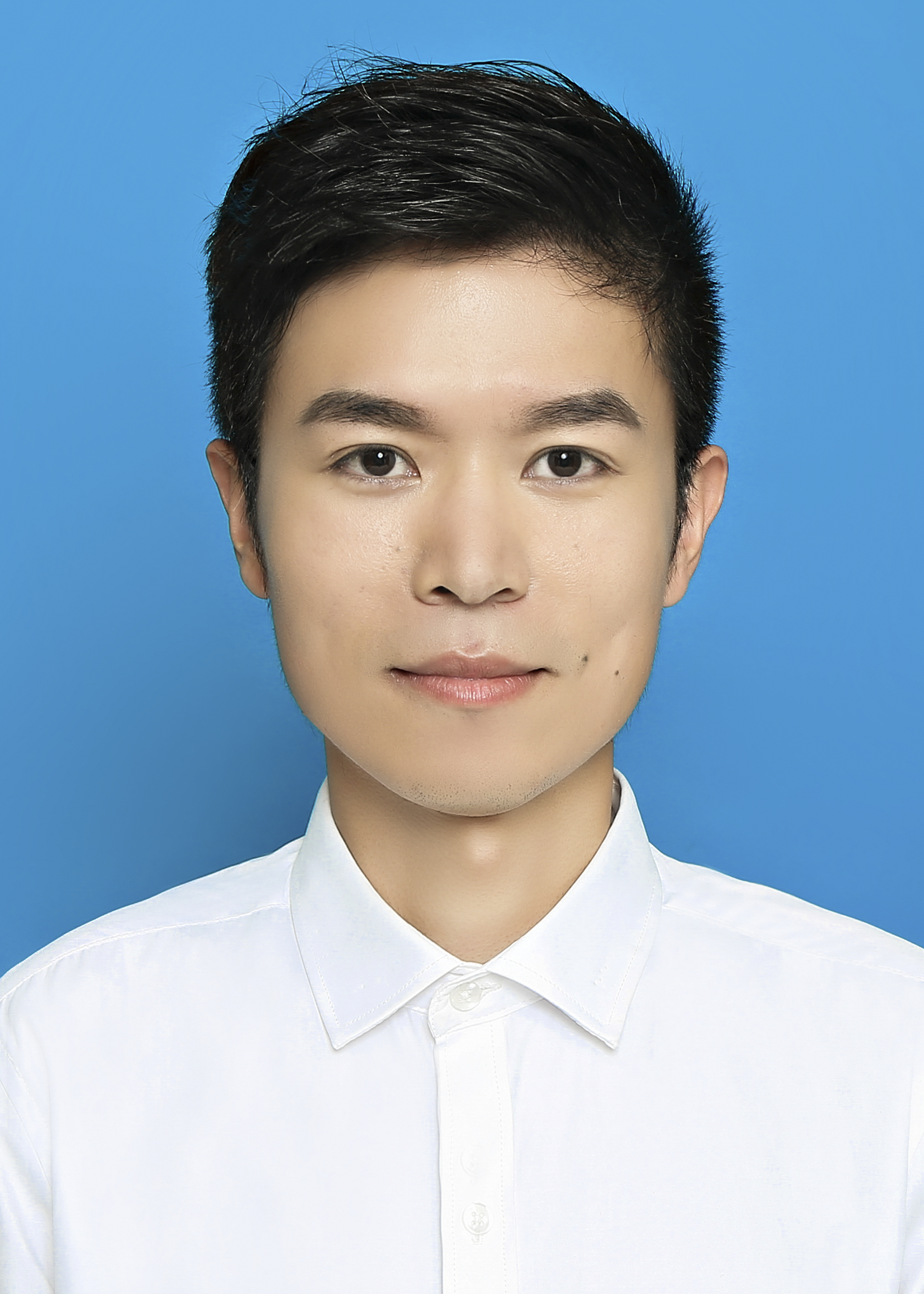}}]
{Chaochao Li} received his Ph.D. degree from the School of Information Engineering, Zhengzhou University, Zhengzhou, China. His current research interests include computer graphics and computer vision. He is currently an assistant research fellow with the School of Computer and Artificial Intelligence, Zhengzhou University, Zhengzhou, China. He has authored over 6 journal and conference papers including the IEEE TRANSACTIONS ON AFFECTIVE COMPUTING, IEEE TRANSACTIONS ON INTELLIGENT TRANSPORTATION SYSTEMS, and IEEE TRANSACTIONS ON SYSTEMS, MAN, AND CYBERNETICS: SYSTEMS.
\end{IEEEbiography}

\begin{IEEEbiography}[{\includegraphics[width=1in,height=1.25in,clip,keepaspectratio]{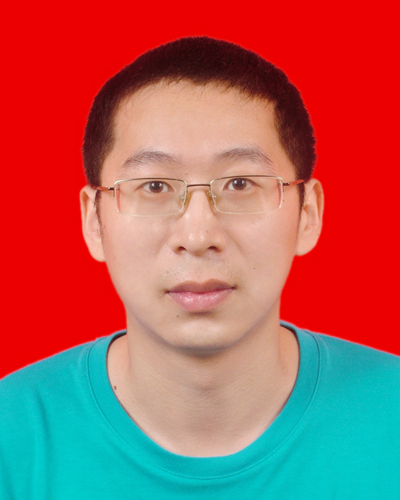}}]
{Mingliang Xu} received the Ph.D. degree in computer science and technology from the State Key Laboratory of CAD\&CG, Zhejiang University, Hangzhou, China, in 2012. He is a Full Professor and the Director with the School of Computer and Artificial Intelligence, Zhengzhou University, Zhengzhou, China. His research interests include computer graphics, multimedia, and artificial intelligence. He has authored more than 100 journal and conference papers in the above areas, including the ACM Transactions on Graphics, the ACM Transactions on Intelligent Systems and Technology, the IEEE TRANSACTIONS ON PATTERN ANALYSIS AND MACHINE INTELLIGENCE, the IEEE TRANSACTIONS ON IMAGE PROCESSING, the IEEE TRANSACTIONS ON CYBERNETICS, the IEEE TRANSACTIONS ON CIRCUITS AND SYSTEMS FOR VIDEO TECHNOLOGY, ACM SIGGRAPH (Asia), ACM MM, and ICCV. 
\end{IEEEbiography}




\end{document}